\documentclass[10pt]{article}
\pdfoutput=1
\PassOptionsToPackage{dvipsnames}{xcolor}

\usepackage[preprint]{tmlr}
\usepackage{algorithm}
\usepackage{algorithmic}
\usepackage{url}

\usepackage{etoolbox}
\makeatletter
\AtBeginEnvironment{algorithmic}{\let\AND\@undefined}
\makeatother

\usepackage{microtype}
\usepackage{graphicx}
\usepackage{subcaption}
\usepackage{booktabs} %

\usepackage{hyperref}

\usepackage{float}
\usepackage[dvipsnames]{xcolor}

\usepackage{mathtools}
\usepackage{amssymb}
\usepackage{amsthm}

\usepackage{subcaption}

\usepackage[capitalize,noabbrev]{cleveref}
\hypersetup{
    colorlinks,
    linkcolor=black,
    citecolor={blue!50!black},
    urlcolor={black}
}

\theoremstyle{plain}
\newtheorem{theorem}{Theorem}[section]

\newtheorem{lemma}[theorem]{Lemma}

\theoremstyle{definition}
\newtheorem{definition}[theorem]{Definition}

\theoremstyle{remark}
\newtheorem{remark}[theorem]{Remark}

\numberwithin{equation}{section}

\makeatletter
\let\c@algorithm\c@theorem
\makeatother

\makeatletter
\let\c@algorithm\c@theorem

\makeatother

\newcommand{\ve}[1]{\boldsymbol{#1}}

\DeclareMathOperator{\vol}{vol}

\DeclareMathOperator{\rank}{rank}

\newcommand{\defeq}{\coloneqq}

\newcommand{\C}{\mathbb{C}}
\newcommand{\R}{\mathbb{R}}
\newcommand{\Q}{\mathbb{Q}}

\newcommand{\N}{\mathbb{N}}

\newcommand{\E}{\mathbf{E}}

\newcommand{\grad}{\nabla}

\DeclareMathOperator{\Rep}{Rep}

\DeclareMathOperator{\GL}{GL}

\title{From Global to Local: A Scalable Benchmark for Local Posterior Sampling}

\author{\name Rohan Hitchcock \email hitchcock@resolution.org \\
      \addr Resolution \\
      University of Melbourne \\
      CSIRO
      \AND
      \name Jesse Hoogland \email jesse@resolution.org \\
      \addr Resolution}

\begin{document}

\maketitle

\begin{abstract}
Degeneracy is an inherent feature of the loss landscape of neural networks, but it is not well understood
how stochastic gradient MCMC (SGMCMC) algorithms interact with this degeneracy. In particular, existing global convergence guarantees 
for common SGMCMC algorithms
rely on assumptions which are likely incompatible with degenerate loss 
landscapes. In this paper, we argue that this gap requires a shift in focus from \emph{global} 
to \emph{local} posterior sampling, and, as a first step, we introduce a novel scalable benchmark 
for evaluating the \emph{local} sampling performance of SGMCMC algorithms. 
We evaluate a number of common algorithms, and find that RMSProp-preconditioned 
SGLD is most effective at faithfully representing the local geometry of the posterior 
distribution among the samplers we evaluate. Although we lack theoretical guarantees about global sampler convergence, 
our empirical results show that we are able to extract non-trivial local information 
in models with up to O(100M) parameters.
\end{abstract}

\section{Introduction}

Neural networks have highly complex loss landscapes which are non-convex and have 
non-unique degenerate minima. When a neural network is used as the basis for a Bayesian 
statistical model, the complex geometry of the loss landscape makes sampling from 
the Bayesian posterior using Markov chain Monte Carlo (MCMC) algorithms difficult. 
Much of the research in this area has focused on whether MCMC algorithms 
can adequately explore the \emph{global} geometry of the loss landscape by visiting 
sufficiently many minima. Comparatively little attention 
has been paid to whether MCMC algorithms adequately explore the \emph{local} geometry 
near minima. Our focus is on Stochastic Gradient MCMC (SGMCMC) algorithms, such as 
Stochastic Gradient Langevin Dynamics (SGLD; \citealt{welling2011SGLD}), applied 
to large models like neural networks. For these models, the local geometry near critical points 
is highly complex and degenerate, so local sampling is a non-trivial problem.

\begin{figure}[h!]
    \centering
    \centerline{\includegraphics[width=\textwidth]{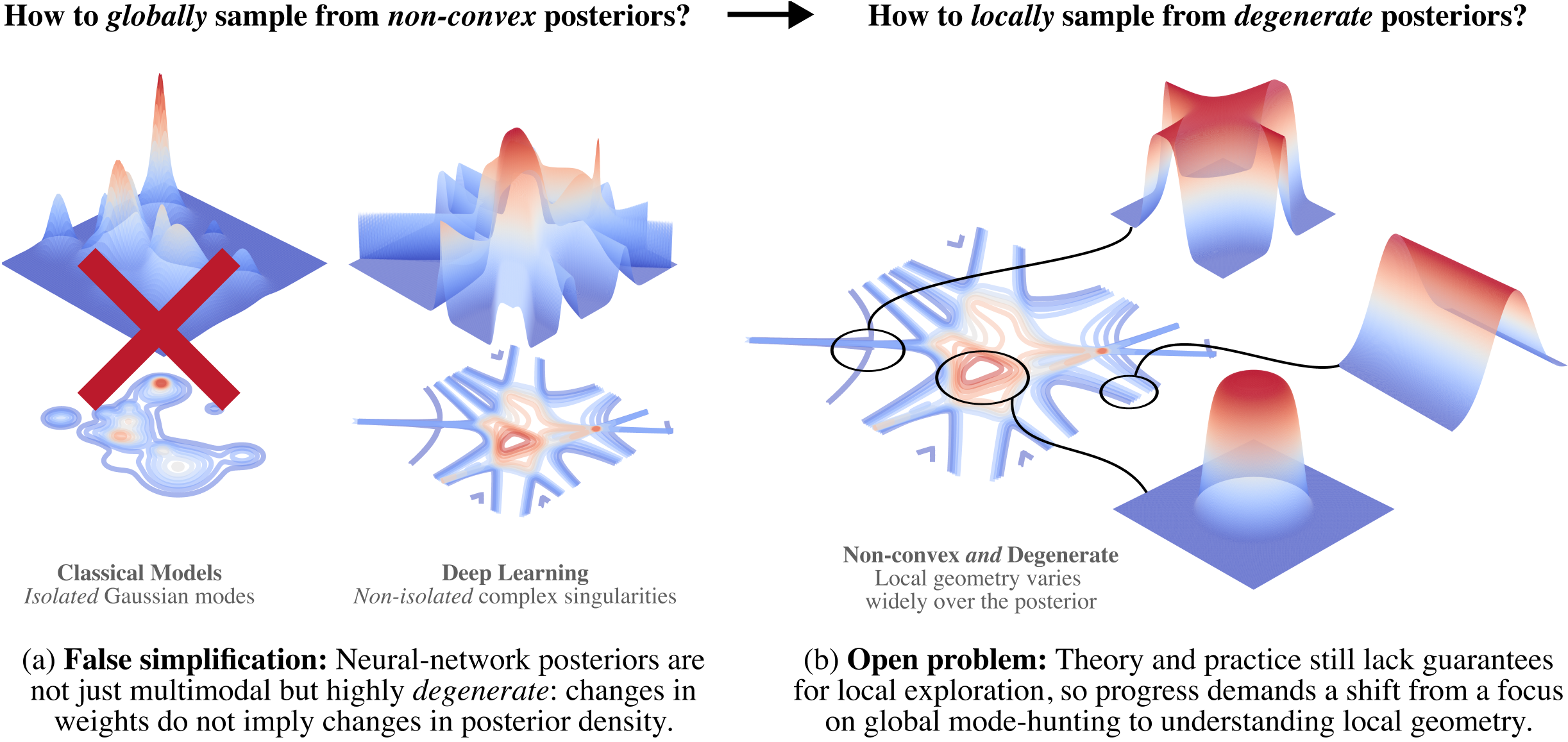}}
    \caption{\textbf{From global to local posterior sampling.} \textit{Left}: Neural network posteriors are often erroneously simplified as isolated Gaussian modes. \textit{Middle}: Neural network posterior distributions are highly \textit{degenerate}, where parameter changes often don’t affect posterior density. \textit{Right}: Local sampling must handle these degeneracies, which raises open theoretical and practical questions about the guarantees and effectiveness of \textit{local} posterior exploration.}
    \label{fig:global-to-local}
\end{figure}

In this paper we argue for a shift in focus from \emph{global} to \emph{local} posterior sampling for complex models like 
neural networks (illustrated in \cref{fig:global-to-local}). Our main contributions are as follows.

\begin{itemize}
    \item \textbf{We identify open theoretical problems regarding the convergence of SGMCMC algorithms for posteriors 
with degenerate loss landscapes.} We survey existing theoretical guarantees for the \emph{global} convergence of SGMCMC algorithms, 
noting their common incompatibility with the degenerate loss landscapes characteristic of these models 
(e.g., deep linear networks). Additionally, we highlight some important negative results which suggest that global 
convergence may in fact \emph{not} occur. Despite this, empirical results show that SGMCMC algorithms are able to extract 
non-trivial \emph{local} information from the posterior; a phenomenon which currently lacks theoretical explanation.
\item \textbf{We introduce a novel, scalable benchmark for evaluating the \emph{local} sampling performance of SGMCMC algorithms.} 
Recognizing the challenge of obtaining theoretical convergence guarantees in degenerate loss landscapes, this benchmark assesses a 
sampler's ability to capture known local geometric invariants related to \emph{volume scaling}. Specifically, it leverages deep linear networks (DLNs), where a key invariant controlling volume scaling 
--- the \emph{local learning coefficient} (LLC; \citealp{lauLocalLearningCoefficient2024}) --- can be computed 
analytically. Our method provides ground-truth for local posterior geometry and, in contrast to \citet{lauLocalLearningCoefficient2024}, allows parameters with varying degrees of local degeneracy to be generated within the same model.
\item \textbf{We find RMSProp-preconditioned SGLD \citep{li2016preconditioned} is most effective at capturing local posterior features 
among the samplers we evaluate.} 
We demonstrate our benchmark on five common SGMCMC algorithms, on models up to O(100M) parameters. This offers 
practical guidance for researchers and practitioners. Importantly, our benchmark shows that although we lack theoretical guarantees 
about global sampling, SGMCMC samplers are able to extract non-trivial local information about the posterior distribution. 
\end{itemize}

\section{Background} 

In \cref{sec:problems-with-global-sampling} we discuss existing theoretical results 
about the convergence of SGLD and related sampling algorithms. We highlight some 
important negative results from the literature, which suggest that global convergence of
algorithms like SGLD are unlikely to occur in loss landscapes with degeneracy, 
and show that existing global convergence guarantees for SGLD rely on assumptions 
that likely do not hold for neural networks. 
In \cref{sec:related-work} we discuss applications and related work.

\subsection{Problems with global sampling} \label{sec:problems-with-global-sampling}

We consider the problem of sampling from an absolutely continuous probability 
distribution $\pi(w)$ on $\R^d$. In Bayesian statistics we often consider the 
tempered posterior distribution
\begin{equation} \label{eqn:tempered-posterior}
    \pi(w) \propto
    \varphi (w) \exp(-n\beta L_n(w))
\end{equation}
where $\{p(x| w)\}_{w\in \R^d}$ is a statistical model, $\varphi(w)$ the prior, 
$D_n = \{X_1, \ldots, X_n \}$ a dataset drawn independently from some distribution $q(x)$, 
$L_n(w) = \tfrac{-1}{n}\sum _{i=1}^n \log p(X_i| w)$ is the empirical negative 
log-likelihood, and $\beta > 0$ is a fixed parameter called the inverse temperature. 
For any distribution, we can consider the \emph{overdamped Langevin diffusion}; 
the stochastic differential equation
\begin{equation} \label{eqn:langevin-diffusion}
    dW_t = \tfrac{1}{2}\grad \log \pi(W_t) dt + dB_t
\end{equation}
where $B_t$ is standard Brownian motion. Under fairly mild assumptions on $\pi(w)$
\citep[see][Theorem 2.1]{robertsTweedieLangevin1996} \eqref{eqn:langevin-diffusion}
has well-behaved solutions and $\pi(w)$ is its stationary distribution. 
The idea of using the \emph{forward Euler-Maruyama discretization} of \eqref{eqn:langevin-diffusion}
to sample from $\pi(w)$ was first proposed by \citet{parisiCorrelationFunctionsComputer1981}
in what is now known as the \emph{Unadjusted Langevin Algorithm} (ULA; 
also known as the Metropolis Langevin Algorithm). For $t=0, 1, \ldots$
we take $w_{t+1} = w_t + \Delta w_t$ where
\begin{equation} \label{eqn:unajusted-langevin-algorithm}
    \Delta w_t = \tfrac{\epsilon}{2} \grad \log \pi(w_t) + \sqrt{\epsilon} \eta _t
\end{equation}
in which $\epsilon > 0$ is the step size and $\eta_0, \eta_1, \ldots$ 
are a sequence of iid standard normal random vectors in $\R^d$. Stochastic 
Gradient Langevin Dynamics (SGLD; \citealt{welling2011SGLD})
is obtained by replacing $\grad \log \pi (w_t)$
in \eqref{eqn:unajusted-langevin-algorithm} with a stochastic estimate 
$g(w_t, U_t)$, where $U_0, U_1, \ldots$ are independent random variables.
When $\pi(w)$ is given by \eqref{eqn:tempered-posterior}, usually $g(w, U)$ is a
mini-batch estimate of the log-likelihood gradient $g(w, U) = -\tfrac{n\beta}{m} \sum _{i=1} ^m \grad \log p(X_{u_i}|w)$
where $U = (u_1, \ldots, u_m)$ selects a random subset of the dataset $D_n$.

\paragraph{Degeneracy can cause samplers to diverge.} 
Issues with ULA when $\log\pi(w)$ has degenerate critical points were first 
noted in \citet[Section 3.2]{robertsTweedieLangevin1996}, where they show that 
ULA will fail to converge to $\pi(w)$ for a certain class of distributions 
when $\log \pi(w)$ is a polynomial with degenerate critical points. 
\citet{mattinglyErgodicitySDEsApproximations2002} relates the convergence of 
forward Euler-Maruyama discretizations of stochastic differential equations 
(and hence the convergence of ULA) 
to a global Lipschitz condition on $\grad \log\pi(w)$, giving examples of distributions 
which do not satisfy the global Lipschitz condition for which ULA diverges at any 
step size. \citet[Theorem 1]{hutzenthalerStrongWeakDivergence2011} shows that if 
$\grad \log\pi(w)$ grows any faster than linearly then ULA diverges. This is a strong 
negative result on the global convergence of ULA, since in models like 
deep linear networks, the presence of degenerate critical points causes super-linear 
growth away from these critical points. 

\paragraph{The current theory makes strong assumptions about the loss landscape.}
Given the above results, it is therefore no surprise that results showing that SGLD is 
well-behaved in a global sense rely either on global Lipschitz conditions on $\grad \log\pi(w)$, 
or on other conditions which control the growth of this gradient. Asymptotic and non-asymptotic 
global convergence properties of SGLD are studied in 
\citet{tehConsistencySGLD2015,vollmerNonAsymptoticProperties2015}. These 
results rely
on the existence of a Lyaponov function $V : \R^d \to [1, \infty)$ with globally bounded 
second derivatives for which there exists $C > 0$ such that
\begin{equation} \label{eqn:teh-et-al-assumption-4-lyaponov}
\|\grad V(w) \|^2 + \| \grad \log \pi (w) \| ^2 \leq C V(w)
\end{equation}
for all $w \in \R^d$ \citep[see][Assumption 4]{tehConsistencySGLD2015}. 
The bounded second derivatives of $V(w)$ impose strong conditions on the growth of $\log \pi (w)$ and 
is incompatible with the deep linear network regression model described in \cref{sec:deep-linear-network-task}. 
A more general approach to analyzing the convergence of diffusion-based 
SGMCMC algorithms is described in \citet{chen2015SGMCMCconvergence}, 
though in the case of SGLD the necessary conditions for this method 
imply that \eqref{eqn:teh-et-al-assumption-4-lyaponov} holds 
\citep[see][Appendix C]{chen2015SGMCMCconvergence}. 

Other results about the global convergence of SGLD and ULA rely on a global 
Lipschitz condition on $\grad \log \pi(w)$ (also called $\alpha$-smoothness), 
which supposes that there exists a constant $\alpha > 0$ such 
that 
\begin{equation} \label{eqn:global-lipschitz-gradient-condition}
    \| \grad \log \pi(w_1) - \grad \log \pi(w_2) \| \leq \alpha \|w_1 - w_2\|
\end{equation}
for all $w_1, w_2 \in \R^d$. This is a common assumption when studying SGLD in the context of stochastic 
optimization \citep{raginskyNonconvexLearningStochastic2017,
tzenLocalOptimalityGeneralization2018,
xuGlobalConvergenceLangevin2020,
zouFasterConvergenceStochastic2021,zhangNonasymptoticEstimatesStochastic2022} and 
also in the study of forward Euler-Maruyama discretizations of diffusion processes 
including ULA
\citep{kushner1987asymptotic,borkarStrongApproximationTheorem1999,durmusNonasymptoticConvergenceAnalysis2017,brossePromisesPitfallsStochastic2018,dalalyanUserfriendlyGuaranteesLangevin2019,chengSharpConvergenceRates2020}. 
Other results rely on assumptions which preclude the possibility of divergence 
\citep{gelfandRecursiveStochasticAlgorithms1991,highamStrongConvergenceEulerType2002}.

Conditions which impose strong global conditions on the growth rate of $\grad \log \pi(w)$ 
often do not hold when $\log \pi(w)$ has degenerate critical points. For concreteness, 
consider the example of an $N$-layer \emph{deep linear network} learning a regression task
(full details are given in \cref{sec:deep-linear-network-task}).
\cref{thrm:dlns-are-high-degree-polynomials} shows that assumptions \eqref{eqn:teh-et-al-assumption-4-lyaponov}
and \eqref{eqn:global-lipschitz-gradient-condition} do not hold when $N > 1$, which is exactly 
the situation when $L_n(w)$ has degenerate critical points.

\begin{lemma} \label{thrm:dlns-are-high-degree-polynomials}
    The negative log-likelihood function $L_n(w)$ for the deep linear network regression task
    described in \cref{sec:deep-linear-network-task} is a polynomial in $w$ of degree $2N$
    with probability one, where $N$ is the number of layers of the network.
\end{lemma}
\begin{proof}
    We give a self-contained statement and proof in \cref{sec:dln-proof}. 
\end{proof}

\paragraph{All is not lost.}

Despite the negative results discussed above and the lack of theoretical guarantees, 
\citet{lauLocalLearningCoefficient2024} shows empirically that SGLD can be used to obtain 
good local measurements of the geometry of $L_n(w)$ when $\pi(w)$ has the form in 
\eqref{eqn:tempered-posterior}. This forms the basis of our benchmark 
in \cref{sec:deep-linear-network-task}, and we show similar results for a variety 
of SGMCMC samplers in \cref{sec:results}. In the absence of theoretical guarantees about 
sampler convergence, we can empirically verify that samplers can recover important geometric 
invariants of the log-likelihood.
We emphasize that this empirical phenomenon is unexplained by the global convergence results 
discussed above, and presents an open theoretical problem. 

Some work  proposes
modifications to ULA or SGLD which aim to address potential convergence issues 
\citep{gelfand1993metropolis,lambaAdaptiveEulerMaruyamaScheme2006,hutzenthalerStrongConvergenceExplicit2012a,
sabanisNoteTamedEuler2013,sabanisHigherOrderLangevin2019,brosseTamedUnadjustedLangevin2019}.
Finally, \citet{zhangHittingTimeAnalysis2018} is notable for its \emph{local} analysis 
of SGLD, studying escape from and convergence to local minima in the context of 
stochastic optimization.

\subsection{The need for local sampling} \label{sec:related-work}

The shift toward local posterior sampling has immediate practical implications in areas such as interpretability and Bayesian deep learning.

\paragraph{Interpretability.}

SGMCMC algorithms play a central role in approaches to interpretability based on 
\emph{singular learning theory} (SLT). SLT
\citep[see][]{watanabeAlgebraicGeometryStatistical2009,watanabeMathematicalTheoryBayesian2018}
is a mathematical theory of the large-data asymptotics of Bayesian learning which properly accounts for 
degeneracy in the model's log-likelihood function, and so is the correct such 
theory for neural networks \citep[see][]{wei2022deepLearningSingular}. 
It provides us with statistically relevant geometric invariants such as the 
\emph{local learning coefficient} (LLC; \citealp{lauLocalLearningCoefficient2024}), 
which has been estimated using SGLD in large neural networks. 
When tracked over training, changes in the LLC correspond to 
qualitative changes in the model's behavior 
\citep{chen2023TMS,hooglandDevelopmentalLandscapeInContext2024,carroll2025EssentialDynamics}. 
This approach has been used in models as large as 100M parameter language models, 
providing empirically useful results for model interpretability \citep{wangDifferentiationSpecializationAttention2024}. 
SGMCMC algorithms are also required to estimate local quantities other than the LLC \citep{baker2025Susceptibilities,gordon_towards_2026,kreerBayesianInfluenceFunctions2025,adamLossKernelGeometric2025}.

\paragraph{Bayesian deep learning.}

Ensembling approaches to Bayesian deep learning involve first training a neural network using a standard 
optimization method, and then sampling from the posterior distribution in a neighborhood of the parameter 
found via standard training. This is done for reasons such as uncertainty quantifications during 
prediction. See \citet{papamarkouPositionBayesianDeep2024} for a survey.

\section{Methodology} \label{sec:benchmarks-for-local-sampling}

In \cref{sec:slt-background} we discuss how the local geometry of 
the expected negative log-likelihood $L(w)$ 
affects the posterior distribution 
\eqref{eqn:tempered-posterior}, focusing on \emph{volume scaling} 
of sublevel sets of $L(w)$. 
In \cref{sec:deep-linear-network-task} we describe a specific benchmark 
for local sampling which involves estimating the local learning 
coefficient of \emph{deep linear networks}. Details of experiments 
are given in \cref{sec:dln-experiments}.

\subsection{Measuring local posterior geometry} \label{sec:slt-background}

In this section we consider the setting of \cref{sec:problems-with-global-sampling}, in particular 
the tempered posterior distribution $\pi(w)$ in \eqref{eqn:tempered-posterior}
and the geometry of the expected negative log-likelihood function 
$L(w) = - \E \log p(X|w)$ where $X\sim q(x)$.

\subsubsection{Volume scaling in the loss landscape.}
An important geometric quantity of $L(w)$ from the perspective of the posterior
$\pi(w)$ is the volume of sublevel sets 
\begin{equation} \label{eqn:pop-loss-sublevel-set}
    V(\epsilon, w_0) = \vol \{ w \in \mathcal{W} \mid L(w) \leq L(w_0) + \epsilon\}
\end{equation}
where $\epsilon > 0$, $w_0$ is a minimum of $L(w)$ and $\mathcal{W} \subseteq \R^d$ is a neighborhood of $w_0$. The 
volume $V(\epsilon, w_0)$ quantifies the number of parameters near $w_0$ which 
achieve close to minimum loss within $\mathcal{W}$, and so is closely related to 
how the posterior distribution $\pi(w)$ concentrates.

\begin{figure}[t]
    \centering
    \centerline{\includegraphics[width=\textwidth]{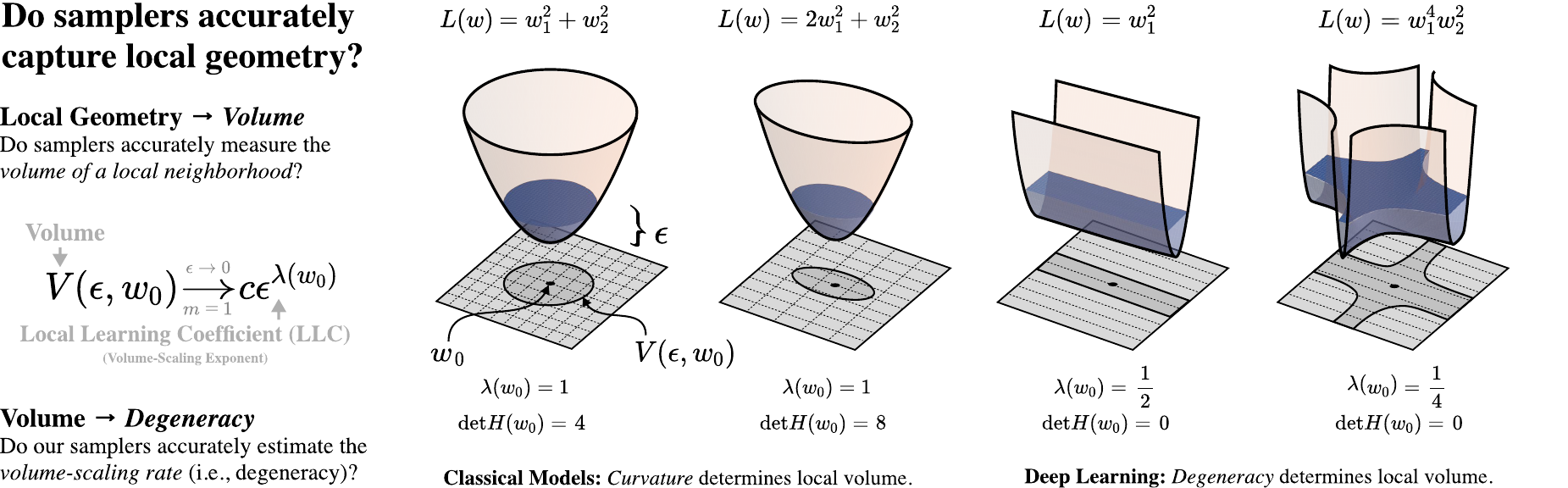}}
    \caption{\textbf{The Local Learning Coefficient (LLC) captures the local geometry of the posterior.}
We illustrate volume-scaling behavior near minima for various simple potentials, highlighting their local geometry. The LLC, defined as the volume-scaling exponent, quantifies the extent of degeneracy. For non-degenerate minima (left two examples), the LLC is always $d/2$, where $d$ is the number of parameters, and the Hessian determinant is nonzero. In contrast, degenerate minima (right two examples) have different LLCs (both less than $d/2$) and a Hessian determinant of $0$, reflecting a geometry fundamentally distinct from Gaussian. Estimating the LLC thus serves as a benchmark for assessing a sampler's capacity to explore complex, degenerate posteriors.}
    \label{fig:llc-volume-scaling}
\end{figure}

We can consider the rate of volume scaling by taking $\epsilon \to 0$. When $w_0$
is a non-degenerate critical point, volume scaling is determined by the spectrum 
of the Hessian $H(w_0)$ at $w_0$ and we have 
\begin{equation} \label{eqn:volume-scaling-regular-case}
    V(\epsilon, w_0) \approx |\det H(w_0)|^{-1/2} \epsilon ^{d/2}  \qquad \text{as}~ 
    \epsilon \to 0
\end{equation}
where $d$ is the dimension of parameter space. However, when $w_0$ is a degenerate critical point 
we have $\det H(w_0) = 0$ and this formula is no longer true; the second-order Taylor expansion
of $L(w)$ at $w_0$ used to derive \eqref{eqn:volume-scaling-regular-case} no longer provides 
sufficient geometric information understand how volume is changing as $\epsilon \to 0$. 
In general we have 
\begin{equation} \label{eqn:volume-scaling-singular-case}
    V(\epsilon, w_0) \approx c\epsilon ^{\lambda (w_0)}(-\log \epsilon)^{m(w_0) - 1}
    \qquad \text{as}~ 
    \epsilon \to 0
\end{equation}
for some $c> 0$, 
where $\lambda(w_0) \in \Q$ is the \emph{local learning coefficient} (LLC) and $m(w_0) \in \N$ is its 
\emph{multiplicity} within $W$ \citep[see][Section 3, Appendix A]{lauLocalLearningCoefficient2024}. 
When $w_0$ is non-degenerate and the only critical point in $\mathcal{W}$ then $\lambda(w_0) = d/2$, 
$c = |\det H(w_0)|^{-1/2}$ and $m(w_0) = 1$, and \eqref{eqn:volume-scaling-regular-case}
is obtained from \eqref{eqn:volume-scaling-singular-case}. \Cref{fig:llc-volume-scaling} illustrates this volume-scaling behavior for several simple potentials, contrasting non-degenerate and degenerate minima.

Although the LLC is an invariant of the population loss $L$ and the 
posterior distribution $\pi(w)$ is built from the empirical loss $L_n$ and prior $\varphi$, 
the LLC still controls how the posterior concentrates when the dataset size $n$ is large. 
This is made precise by singular learning theory \citep{watanabeAlgebraicGeometryStatistical2009,watanabeMathematicalTheoryBayesian2018}, 
which says that as $n$ grows the posterior concentrates in regions determined (to leading order) 
by the LLC \citep[see also][]{lauLocalLearningCoefficient2024,chen2023TMS}.

\subsubsection{Measuring volume scaling via sampling.}
A SGMCMC algorithm which is producing good posterior samples from a neighborhood of 
$w_0$ should reflect the correct volume scaling rate $\lambda (w_0)$ in 
\eqref{eqn:volume-scaling-singular-case}. In other words, it should produce good estimates
of the LLC. The LLC can be estimated by sampling from the posterior 
distribution \eqref{eqn:tempered-posterior} without direct access to $L(w)$.

\begin{definition}[{\citealp[Definition 1]{lauLocalLearningCoefficient2024}}] \label{def:llc-estimator}
    Let $w_0$ be a local minimum of $L(w)$ and let $\mathcal{W}$ be an open, connected neighborhood 
    of $w_0$ such that (a) the closure $\overline{\mathcal{W}}$ is compact, (b)
    $L(w_0) = \inf _{w\in \overline{\mathcal{W}}} L(w)$, and (c) $\lambda(w_0) \leq \lambda(w)$
    for any $w \in \overline{\mathcal{W}}$ which satisfies $L(w) = L(w_0)$. 
    The \emph{local learning coefficient estimator} $\hat{\lambda} (w_0)$
    at $w_0$ is
    \begin{equation} \label{eqn:llc-estimator}
        \hat{\lambda}(w_0) = n\beta (\E^{\beta}_w [L_n(w) ] - L_n(w_0) ) 
    \end{equation}
    where $\E^{\beta} _w$ is an expectation over the tempered posterior \eqref{eqn:tempered-posterior} with prior $\varphi(w)$ and support $\overline{\mathcal{W}}$.
\end{definition}

The following theorem relies on a number of technical 
conditions, which we give in \cref{def:fundamental-slt-conditions} in \cref{sec:slt-technical-conditions-appendix}.

\begin{theorem}[{\citealp[Theorem 4]{watanabe2013WABIC}}] \label{thrm:llc-wbic-estimator}
    Let $w_0$ be a local minimum of $L(w)$ and consider the local learning coefficient estimator $\hat \lambda(w_0)$. 
    Let the inverse temperature in \eqref{eqn:tempered-posterior} be $\beta = \beta_0 / \log n$ for some 
    $\beta_0 > 0$. Assuming the fundamental conditions of SLT given in \cref{def:fundamental-slt-conditions}, 
    $\hat{\lambda}(w_0)$ is an asymptotically unbiased estimator of $\lambda(w_0)$ as $n\to\infty$. 
\end{theorem}

In \citet{lauLocalLearningCoefficient2024}, locality of $\hat{\lambda}(w_0)$ is enforced by using a Gaussian prior 
$\varphi(w) \propto e^{-\gamma \|w - w_0 \|^2/2 }$ centered at $w_0$, 
where $\gamma > 0$ is a hyperparameter of the estimator. We use the same prior in our experiments, 
acknowledging that this deviates from the theory above because it is not compactly supported. The way we 
use the prior in each sampling algorithm is made explicit in the pseudocode in \cref{sec:samplers}.

\subsection{Deep linear network benchmark} \label{sec:deep-linear-network-task}

As noted in \cref{sec:problems-with-global-sampling}, we lack theoretical convergence guarantees for 
SGMCMC algorithms in models such as neural networks. In the absence of these theoretical guarantees, 
we can instead empirically verify that samplers respect certain geometric invariants of the log-likelihood
function. The local learning coefficient (LLC) from \cref{sec:slt-background} is a
natural choice: in contrast to predictive metrics, which assess how well a model's predictions match
the data-generating process, recovering the LLC measures how faithfully the local posterior geometry
itself is represented.

We do not have ground-truth values for the LLC for most systems; 
the only known method for computing it exactly (i.e., other than via statistical 
estimators) involves computing a \emph{resolution of singularities}
\citep{hironakaResolutionSingularitiesAlgebraic1964}, making the problem intractable in general. 
However, LLC values have recently been computed for
\emph{deep linear networks} (DLNs; \citealp{aoyagiConsiderationLearningEfficiency2024}). 
DLNs provide a scalable setting where the ground-truth LLC values are known.

\begin{definition} \label{def:deep-linear-network}
    A \emph{deep linear network} (DLN) with $N$ layers of sizes $d_0, \ldots, d_N$ is a family
    of functions $F (-; \ve W) : \R^{d_0} \to \R^{d_N}$ parametrized by vectors of matrices 
    $\ve W = (W_1, \ldots, W_N)$ where $W_i$ is a $d_i \times d_{i-1}$ matrix. We define 
    $f(x; \ve W) = W x$
    where $W = W_N W_{N-1} \cdots W_1$.
\end{definition}

We fix integers $N$ and $d_0, \ldots, d_N$ for the number of layers and layer widths of 
a DLN architecture. To apply the theoretical results that give the ground-truth 
LLC, we take $d_0 = d_1 = \cdots = d_N$ in all experiments.

The learning task takes the form of a regression task using the parametrized family of 
DLN functions defined in \cref{def:deep-linear-network}, with the 
aim being to learn a function $H : \R^{d_0} \to \R^{d_N}$ given by $H(x) = Bx$
for some $d_N \times d_0$ matrix $B$. We fix a prior $\varphi(\ve W)$
on the set of all matrices parametrizing the DLN.
Consider an input distribution $q(x)$ on $\R^{d_0}$ and a fixed input dataset 
$X_1, \ldots, X_n$ drawn independently from $q(x)$. For a parameter $\ve W$, we 
consider noisy observations of the function's behavior $F(X_i ; \ve W) + Z_i$, where
$Z_i \sim \mathcal{N}(0, \sigma ^2 I_{d_N})$ are independent sources of noise.
The likelihood of observing an input-output pair $(x, y)$ generated using $q(x)$ and the parameter 
$\ve W$ is
\begin{equation} \label{eqn:model-regression-task}
    p(x, y|\ve W) \propto \exp\left(-\frac{1}{2\sigma^2} \|y - F(x;\ve W)\|^2\right) q(x) .
\end{equation}
Likewise, we consider the true distribution 
\[
    q(x, y) \propto \exp\left(-\frac{1}{2\sigma^2} \|y - H(x)\|^2\right) q(x)
\]
on input-output pairs generated by the target function $H$. 
The dataset consists of pairs $(X_i, Y_i)$ where 
$Y_i = G(X_i) + Z_i$. We define
\begin{equation} \label{eqn:expected-negative-log-likelihood-regression-task}
    L(\ve W) = \int _{\R^{d_0}} \| f(x; \ve W) - H(x) \| ^2 q(x) dx
\end{equation}
which is, up to an additive constant, the expected negative log-likelihood $-\E \left[p(X, Y | \ve W) \right]$
where $(X, Y) \sim q(x, y)$. This can be estimated using the dataset as 
\begin{equation} \label{eqn:empirical-negative-log-likelihood-regression-task}
    L_n(\ve W) = \frac{1}{n} \sum _{i=1} ^n \| Y_i - F(X_i; \ve W)\| ^2 .
\end{equation}
Let 
\begin{equation} \label{eqn:dln-optimal-parameters}
    \mathcal{W}_0 = \left\{\ve W \mid L(\ve W) = 0\right\} 
\end{equation}
denote the set of optimal DLN parameters for the above regression task.

\subsubsection{Generating parameters with known local learning coefficients} \label{sec:generating-parameters}

While a DLN with $N\geq 2$ layers expresses exactly the same functions 
as a DLN with $N=1$ layer, the geometry of $\mathcal{W}_0$ defined 
in \eqref{eqn:dln-optimal-parameters}
is significantly more complex. When $N = 1$ the set $\mathcal{W}_0$ consists of a 
single isolated and non-degenerate minimum of $L(\ve W)$, while when $N\geq 2$ it 
is an algebraic variety with non-trivial geometry.
Theoretical results of 
\citet{aoyagiConsiderationLearningEfficiency2024, lehalleurGeometryFibersMultiplication2024}
allow us to study different kinds of points in $\mathcal{W}_0$ and obtain ground-truth values 
for the local learning coefficient at these points. We consider two kinds of parameters:
\begin{itemize}
    \item \textbf{Special optimal parameters.} We generate parameters $\ve W \in 
\mathcal{W}_0$ for which the LLC is minimal over $\mathcal{W}_0$. Because the LLC is minimal, it
can be computed using the formula of \citet[Theorem 1]{aoyagiConsiderationLearningEfficiency2024},
which we restate as \cref{thrm:aoyagi-dln-apdx}.
    \item \textbf{Generic optimal parameters.} We generate an arbitrary parameter on 
$\ve W \in \mathcal{W}_0$ in such a way that we have the necessary combinatorial data 
to compute the LLC at $\ve W$. The LLC can be computed by solving a sequence of constrained 
integer optimization problems. This approach builds on the results of \citet{lehalleurGeometryFibersMultiplication2024}.
\end{itemize}

From the perspective of evaluating sampling algorithms, the special and generic parameters allow 
us to evaluate how a sampler performs for different kinds of local degeneracy within
the same model. It also improves upon the method in \citet{lauLocalLearningCoefficient2024}, which 
studies LLC estimation using SGLD. There, DLN parameters are randomly generated and the estimated 
LLC is compared with the minimum LLC value of \citet[Theorem 1]{aoyagiConsiderationLearningEfficiency2024}. 
This provides a lower bound on the ground truth, while in our method we know the ground truth value exactly.

The geometry of $\mathcal{W}_0$ is studied in \citet{lehalleurGeometryFibersMultiplication2024}
via a group action on parameter space. 
Let $\GL_{d_i}$ be the group of invertible $d_i \times d_i$ matrices and let
$G = \prod_{i=0}^N \GL_{d_i}$ act on a parameter $\ve W = (W_1, \ldots, W_N)$ by change of basis:
\begin{equation} \label{eqn:group-action-main}
    g \cdot \ve W = (P_1 W_1 P_0^{-1}, P_2 W_2 P_1^{-1}, \ldots, P_N W_N P_{N-1}^{-1}),
    \qquad g = (P_0, \ldots, P_N) \in G .
\end{equation}
Recall that the \emph{$G$-orbit} of a parameter $\ve W$ is the set $\{g \cdot \ve W \mid g\in G\}$. 
\citet{lehalleurGeometryFibersMultiplication2024} exploit a relationship between the set 
of all $G$-orbits and two different kinds of combinatorial objects: \emph{Kostant partitions}
and \emph{rank patterns} (see \cref{def:kostant-partition-rank-pattern}). The sets of all 
$G$-orbits, Kostant partitions, and rank patterns are related via explicitly given bijections 
(see \cref{lemma:orbit-bijection}).

The geometry of $\mathcal{W}_0$ is organized by a partial order on rank patterns, defined in \eqref{eqn:rank-pattern-partial-order}
in \cref{sec:background-geometry-of-dlns}. 
The special parameters are exactly the orbits of \emph{minimal} rank patterns, which \citet{lehalleurGeometryFibersMultiplication2024}
show realize the minimal LLC (see \cref{thrm:special-true-parameter-characterisation}). In this 
case the LLC can be computed using \cref{thrm:aoyagi-dln-apdx}.
For a generic parameter the LLC is instead recovered from its rank pattern by maximizing over this partial order
(see \cref{thrm:llc-from-orbits}). Computing this involves solving a sequence of constrained integer-optimization problems
and is described in detail in \cref{sec:appendix-computing-llc}. 

Additional detail is given in \cref{sec:geometry-of-dlns}: in \cref{sec:background-geometry-of-dlns}
we discuss the approach and results of \citet{aoyagiConsiderationLearningEfficiency2024,lehalleurGeometryFibersMultiplication2024}, 
in \cref{sec:generating-special-and-generic-params,sec:appendix-computing-llc} we describe our method 
for generating parameters and computing LLCs
in more detail. \cref{alg:generate-special,alg:generate-generic,alg:find-maximal-rank-patterns}
give pseudocode for generating special parameters, generic parameters, and computing the LLCs of generic
parameters respectively. As part of generating both special and generic parameters, we generate 
random elements of $G$. This requires some thought to ensure that the resulting DLN parameters are 
numerically well-behaved; we describe our method in \cref{sec:generating-special-and-generic-params}
and \cref{alg:generate-g}.

\subsection{Deep linear network experiments} \label{sec:dln-experiments}

We generate special and generic DLN parameters as described in \cref{sec:generating-parameters}
and \cref{sec:generating-special-and-generic-params}.
For the special parameters, we consider four classes of DLN architecture 
100K, 1M, 10M and 100M, whose names approximately correspond to the total number of 
parameters of the DLN. 
For generic parameters, we consider a class of DLN architecture with up to approximately 
10K parameters. We need to use smaller DLNs for experiments involving generic parameters 
as computing the local learning coefficient in this case involves solving a sequence of 
constrained integer optimization problems (see \cref{alg:find-maximal-rank-patterns}) 
which become computationally infeasible as the size of the DLN grows. 
The number of layers and width for all DLN architecture classes is generated according to the
values in \cref{tab:dln-architecture-hparams} in \cref{sec:generating-special-and-generic-params}.

We estimate the LLC by running a given SGMCMC algorithm for $T$ steps starting 
at $\ve W_0$. Following \citet{lauLocalLearningCoefficient2024}, we enforce locality using a Gaussian 
prior centered at $\ve w_0$.
We provide pseudocode for our implementation in \cref{sec:samplers}.

This results in a sequence $\ve W_0, \ve W_1, \ldots, \ve W_T$ of parameters. For all SGMCMC algorithms we use 
mini-batch estimates $L_{m, t}(\ve W_t)$ of $L_n(\ve W_t)$, where
$L_{m,t}(\ve W_t) = \tfrac{1}{m}\sum \nolimits _{j=1} ^m  \| f(X_{u_{j, t}}; \ve W_t) - Y_{u_{j, t}} \|^2$
and $U_t = (u_{1, t}, \ldots, u_{m, t})$ defines $t$-th batch of the dataset. 
Rather than using $L_n(\ve W_t)$ to estimate the LLC, instead 
assume that 
\[
\overline{L} \approx \E^{\beta} _{\ve W} L_n(\ve W)
\qquad\text{where}\qquad 
\overline{L} = \frac{1}{T - B} \sum \nolimits _{t=B} ^T L_{m, t}(\ve W_t)
\]
for some number of `burn-in' steps $B$. 
Inspired by \cref{def:llc-estimator}, 
we then estimate the LLC at $\ve W_0$ as 
\begin{equation}
    \overline{\lambda}(\ve W_0) \defeq n\beta (\overline{L} - L_{m, 0}(\ve W_0) ) . 
\end{equation}
To improve this estimate, one could run $C > 1$ independent sampling chains to obtain estimates 
however in this case we take $C = 1$, preferring 
instead to run more experiments with different architectures. We give the hyperparameters used in 
LLC estimation in \cref{tab:llc-estimation-hyperparameters} in \cref{sec:deep-linear-network-appendix}.

\section{Results} \label{sec:results}

\begin{figure}[t]
    \centering
    \centerline{\includegraphics[width=\textwidth]{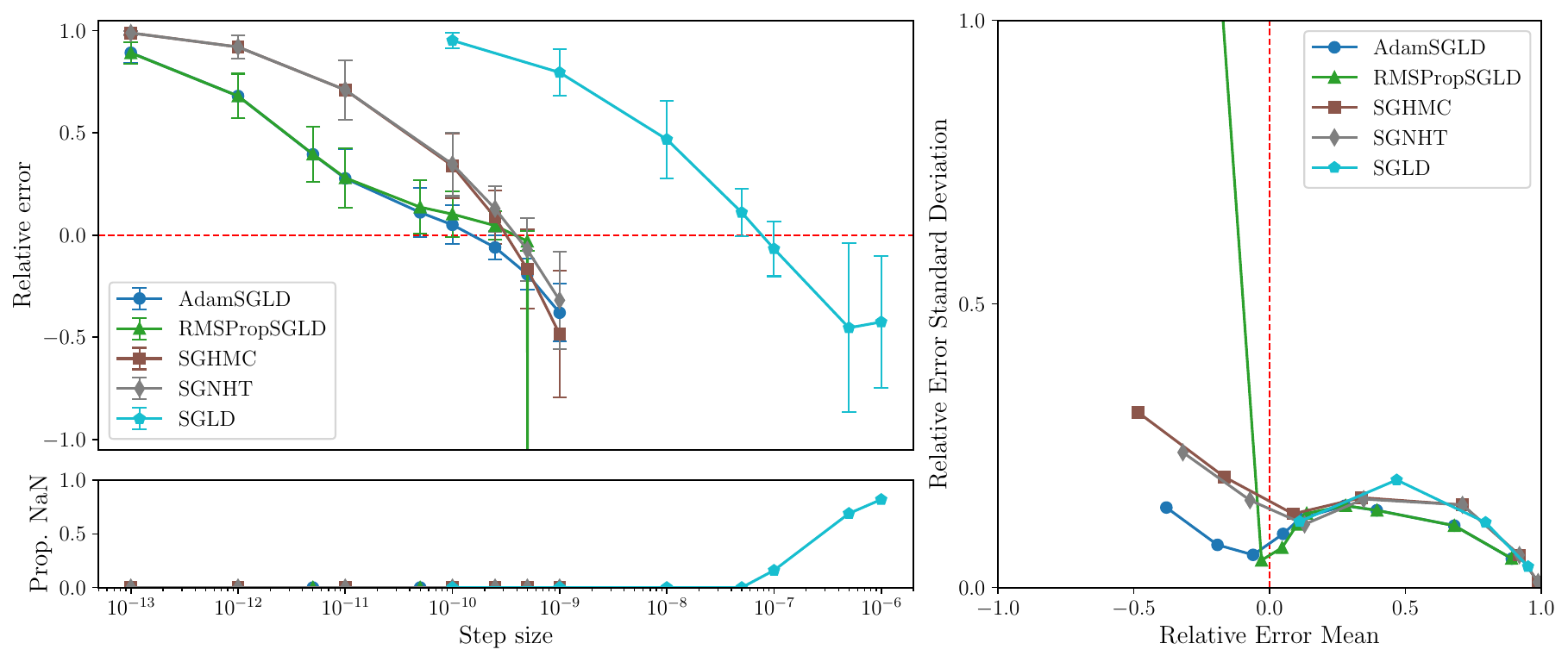}}
    \caption{\textbf{Adaptive samplers like RMSPropSGLD and AdamSGLD achieve 
    superior performance in estimating the local learning coefficient.} 
    These results are for the most degenerate special DLN parameters in 10M parameter models. 
    \textit{Top left:} The mean relative error $(\lambda - \hat{\lambda}) / \lambda$
    versus the step size of the sampler; bars indicate standard deviation. \textit{Bottom left:} The proportion of 
    estimated values which were NaN, indicating the sampler encountered numerical issues. 
    RMSPropSGLD and AdamSGLD are less sensitive to step size, producing more 
    accurate results across a larger range of step size values. \textit{Right:} The mean relative error versus the standard deviation of the 
    relative error, only plotting points where $<10\%$ of estimates are NaN. RMSPropSGLD and AdamSGLD
    achieve a superior mean-variance trade-off. We show the 10M model class here as a representative example; results for all model classes (100K--100M) are given in \cref{sec:deep-linear-network-appendix-additional-results}.}
    \label{fig:relative-error-results-10M}
\end{figure}

To assess how well they capture the local posterior geometry, we apply the 
benchmark described in \cref{sec:deep-linear-network-task} to the 
following samplers: SGLD \citep{welling2011SGLD}, SGLD with RMSProp preconditioning (RMSPropSGLD; \citealt{li2016preconditioned}), an ``Adam-like'' adaptive SGLD (AdamSGLD; \citealt{kim2020AdamSGLD}), 
SGHMC \citep{chenSGHMC2014} and SGNHT \citep{ding2014SGNHT}. We give pseudocode for 
our implementation these samplers in \cref{sec:samplers}. We are interested not only in the
absolute performance of a given sampler, but also in how sensitive its estimates are to the chosen step size
$\epsilon$ of the sampler. To assess 
the performance of a sampler at a given step size, we primarily consider the \emph{relative error}
$(\lambda - \hat{\lambda}) / \lambda$ of an LLC estimate $\hat \lambda$ with true value $\lambda$. 

\paragraph{RMSPropSGLD and AdamSGLD are less sensitive to step-size.}

\Cref{fig:relative-error-results-10M} displays the mean relative error (averaged over different networks 
generated from the model class) versus the step size of each sampler. While all samplers seem to 
be able to achieve empirically unbiased LLC estimates (relative error = 0) for \emph{some} step size, RMSPropSGLD and 
AdamSGLD have a wider range of step size values which produce accurate results. 
For SGLD, at the step size values where the relative error is close to zero, a significant fraction of the 
LLC estimates are also diverging. The same pattern holds across the other model classes (\cref{sec:deep-linear-network-appendix-additional-results}).

\paragraph{RMSPropSGLD and AdamSGLD achieve a superior mean-variance tradeoff.}

\Cref{fig:relative-error-results-10M} (right) plots the mean relative error against its standard deviation 
for each sampler at different step sizes: RMSPropSGLD and AdamSGLD obtain a superior combination of good mean performance and lower variance compared to the other samplers, again consistently across model classes (\cref{sec:deep-linear-network-appendix-additional-results}).

\paragraph{RMSPropSGLD and AdamSGLD are better at preserving order.}

In many applications of LLC estimation, observing relative changes (e.g. over training) in the LLC is more important than determining absolute values \citep{chen2023TMS,hooglandDevelopmentalLandscapeInContext2024,wangDifferentiationSpecializationAttention2024}. 
In such cases, the order of a sampler's estimates should reflect the order of the true 
LLCs. We compute the \emph{order preservation rate} of each sampler, 
which we define as the proportion of all pairs of estimates for which the order of the true LLCs
matches the estimates. 
Again, RMSPropSGLD and AdamSGLD achieve superior performance 
to the other samplers with a higher order preservation rate across a wider range of step sizes; see \cref{fig:order-preservation-rate} in \cref{sec:deep-linear-network-appendix-additional-results}.

\paragraph{RMSPropSGLD step size is easier to tune.}

Above a certain step size RMSPropSGLD experiences rapid performance degradation, with LLC 
estimates which are orders of magnitude larger than the maximum theoretical value of $d/2$
and large spikes in the loss trace of the sampler. In contrast, AdamSGLD experiences 
a more gradual performance degradation and its loss traces do not obviously suggest the 
step size is set to high. We see this catastrophic drop-off in performance as an 
\emph{advantage} of RMSPropSGLD, as it provides a clear signal that hyperparameters are 
incorrectly tuned in the absence of ground-truth LLC values. This clear signal is 
not present for AdamSGLD.

\begin{figure}
    \centering
    \centerline{\includegraphics[width=\textwidth]{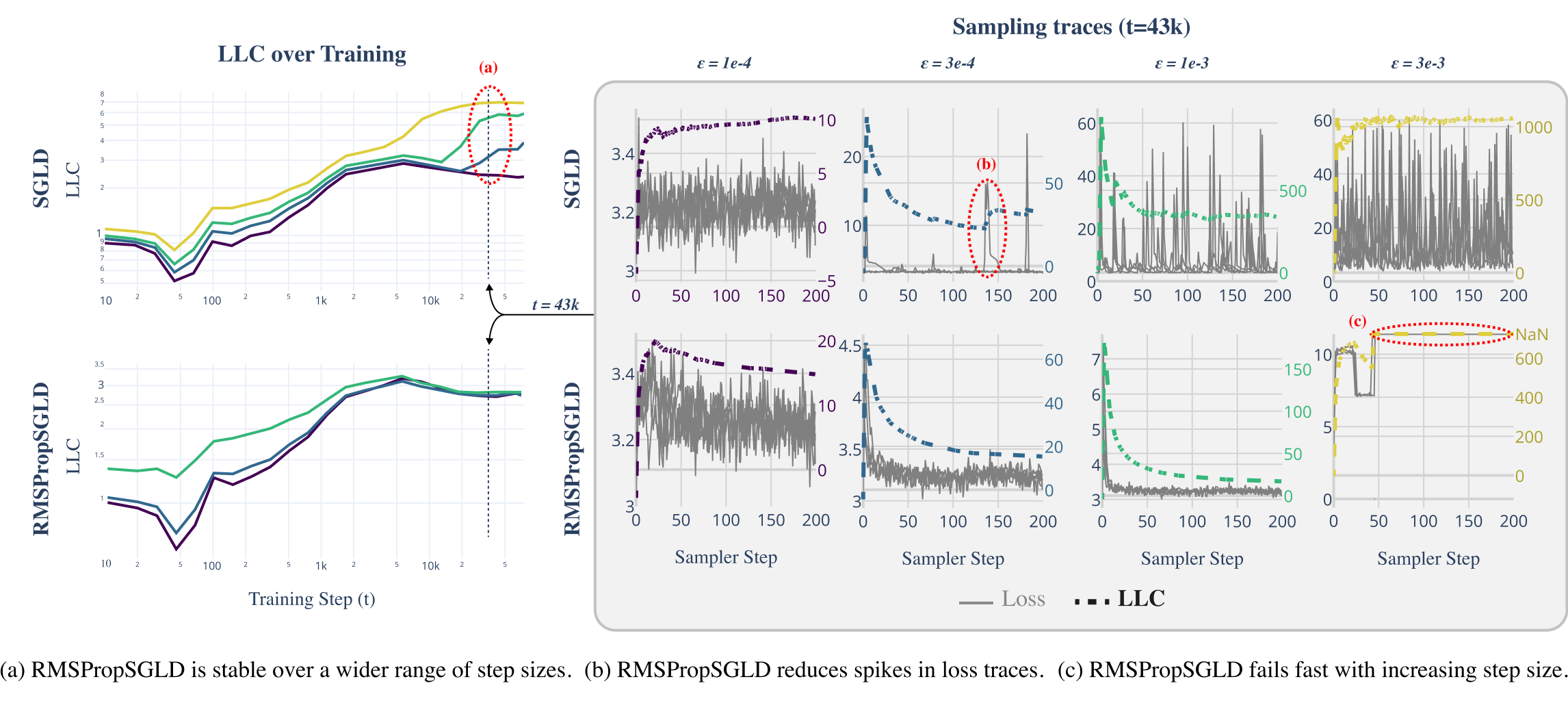}}
    \caption{\textbf{RMSPropSGLD stabilizes sampling chains} for an attention head in a four-layer attention-only transformer trained on the Pile (\cref{sec:llm-experiments}), leading to more consistent and reliable LLC estimates. %
    }
    \label{fig:llm-rmsprop-vs-sgld}
\end{figure}

\paragraph{The advantages of RMSPropSGLD carry over to a real language model.}
To test whether these benchmark findings generalize beyond deep linear networks, we estimate the
local learning coefficient of individual attention heads in a four-layer attention-only transformer
trained on the Pile \citep{gaoPile800GBDataset2020a,xie2023data}, tracking the (weight-refined) LLC
over training \citep{wangDifferentiationSpecializationAttention2024}. Here there is no ground-truth
LLC, so we compare samplers by the stability and consistency of their estimates rather than by
accuracy. RMSPropSGLD produces consistent LLC-over-training while SGLD does not, and the 
loss traces for RMSPropSGLD show fewer spikes
(\cref{fig:llm-rmsprop-vs-sgld}). When the step size is too large RMSPropSGLD fails conspicuously,
returning NaN values, whereas SGLD can return plausible but inaccurate estimates with no obvious
warning sign. Full experimental
details are given in \cref{sec:llm-experiments}.

\paragraph{Sampler behavior is significantly affected by degeneracy.}

We compare the performance of each sampler at both generic (less degenerate) and special (more degenerate) 
parameters and find the samplers perform differently at each type of parameter. The LLC estimates 
produced by each sampler have higher variance at special parameters compared to its performance
at generic parameters. Furthermore, the samplers respond differently to the change in degeneracy. For example, 
SGLD consistently underestimates the LLC at special parameters but is accurate at generic parameters, 
while AdamSGLD and RMSPropSGLD systematically overestimate the LLC at generic parameters but achieve 
a smaller variance. We discuss this in more detail in \cref{sec:generic-vs-special-results}, 
\cref{fig:adamsgld-generic-vs-special,fig:rmsprop-generic-vs-special,fig:sghmc-generic-vs-special,fig:sgnht-generic-vs-special,fig:sgld-generic-vs-special}.

\section{Discussion} \label{sec:discussion}

In this paper we introduced a scalable benchmark for evaluating the \textit{local} sampling 
performance of SGMCMC algorithms. This benchmark is based on how well samplers 
estimate the \emph{local learning coefficient} (LLC) for deep linear networks (DLNs) at 
different types of points in the set of optimal parameter. Since the LLC 
is the local volume-scaling rate for the log-likelihood function, 
this directly assesses how well samplers explore the local posterior. %

\paragraph{The benchmark as a testbed for sampler development.}
Because the LLC provides ground-truth local geometry, the LLC estimation benchmark is a natural testbed for developing and
comparing SGMCMC algorithms. The samplers we study are not
exhaustive, and our finding that both RMSProp preconditioning and the momentum-based SGHMC improve on
plain SGLD suggests that combining such techniques is a promising direction. Preconditioned and
scale-adapted variants of SGHMC \citep{ma_complete_2015,springenberg_bayesian_2016}, and adaptive
schemes more broadly, are natural candidates to evaluate in this way.

\paragraph{Towards future empirical benchmarks.} The stochastic optimization literature has accumulated numerous benchmarks for assessing (optimization) performance on non-convex landscapes (e.g., Rastrigin, Ackley, and Griewank functions; \citealt{plevris2022collection}). However, these benchmarks focus primarily on multimodality and often ignore degeneracy. Our current work takes DLNs as an initial step towards developing a more representative, degeneracy-aware benchmark. A key limitation is that DLNs represent only one class of degenerate models and may not capture all forms of degeneracy encountered in general \citep[see][]{lehalleurGeometryFibersMultiplication2024}. Developing a wider set of degeneracy-aware benchmarks therefore remains an important direction for future research.

\paragraph{Towards future theoretical guarantees.}
In~\cref{sec:problems-with-global-sampling}, we establish that global convergence guarantees for 
sampling algorithms like SGLD rely on assumptions which \emph{provably} do not 
hold for certain model classes (e.g. deep linear networks) with degenerate loss landscapes, 
and are unlikely to be compatible with degeneracy in general. 
Shifting from \emph{global} to \emph{local} convergence guarantees, which properly account for degeneracy, provides one promising way forward. 

This shift may have broader implications beyond sampling. Many current convergence guarantees for stochastic optimizers make similar assumptions that may fail for degenerate landscapes (e.g., global Lipschitz  or Polyak-Łojasiewicz conditions; \citealt{rebjockFastConvergenceNonisolated2024}). Generally, the role of degeneracy in shaping the dynamics of sampling and optimization methods is not well understood.

\paragraph{Open problem: A theoretical explanation for the empirical success of \textit{local} SGMCMC.}
In this paper, we observed empirically that SGMCMC algorithms can successfully estimate the LLC despite the lack of theoretical convergence guarantees. The strong assumptions employed 
by the convergence results discussed in \cref{sec:problems-with-global-sampling} arise, in some sense, because 
the goal is to prove \emph{global} convergence to a posterior with support $\R^d$. It possible is that convergence 
results similar to those in \cref{sec:problems-with-global-sampling} may be proved for compact parameter spaces, 
however this does not explain the success of local sampling we observe because our experiments are not in 
a compactly supported setting.
This \emph{also} places 
our empirical results outside of the setting of SLT (in particular
\cref{thrm:llc-wbic-estimator}). We see understanding precisely 
what determines the ``effective support'' of SGMCMC sampling chains in practice as a central issue 
in explaining why these samplers work in practice.

\makeatletter\if@accepted
\subsubsection*{Acknowledgments}
We are grateful to Simon Pepin Lehalleur for patiently explaining the results of
\citet{lehalleurGeometryFibersMultiplication2024}, and for feedback on an earlier version
of this work, which led to our treatment of generic and special optimal parameters.
This work was completed while Rohan Hitchcock was a PhD candidate at the University of
Melbourne and CSIRO. Rohan Hitchcock was supported by CSIRO through the AI for Missions
PhD Program, and by the Commonwealth through an Australian Government Research Training
Program Scholarship (\url{https://doi.org/10.82133/C42F-K220}). Computational resources
used in this work were provided by CSIRO.
\fi\makeatother

\bibliographystyle{tmlr}
\bibliography{main}

\newpage 
\appendix 

\section*{Appendix}

This appendix contains supplementary details, proofs, and experimental results supporting the main text. Specifically, we include:

\begin{itemize}
    \item \cref{sec:examples-of-degeneracy} provides examples of global and local degeneracies that are characteristic of modern deep neural network architectures. 
    \item \cref{sec:geometry-of-dlns} gives the geometric background and full theorem statements underlying the parameter-generation method of \cref{sec:generating-parameters}, together with the integer optimization problem used for generic parameters.
    \item \cref{sec:dln-proof} provides a proof of \cref{thrm:dlns-are-high-degree-polynomials}, which shows that the negative log-likelihood of deep linear networks is a polynomial of degree $2N$.
    \item \cref{sec:deep-linear-network-appendix} presents additional experimental results for deep linear networks. In particular, 
    we discuss the differences in sampler behavior at special and generic parameters in \cref{sec:generic-vs-special-results}.
    \item \cref{sec:llm-experiments} describes additional methodological details for the LLM experiments. 
    \item \cref{sec:slt-technical-conditions-appendix} summarizes the fundamental technical conditions required by singular learning theory (SLT), outlining the mathematical assumptions underlying our theoretical discussions.
    \item \cref{sec:samplers} gives pseudocode for the specific sampler implementations we use in experiments. 
\end{itemize}

\newpage

\section{Examples of Degeneracy}\label{sec:examples-of-degeneracy}

Degenerate critical points in the loss landscape of neural networks can arise 
from \emph{symmetries} in the parametrization: continuous (or discrete) families of parameter settings
that induce \emph{identical} model outputs or leave the training loss unchanged.
We distinguish \textit{global} (or ``generic'') symmetries, which hold throughout parameter space,
from \textit{local} and \textit{data-dependent} degeneracies that arise only in particular regions in parameter space or only for particular data distributions. In this section we provide several examples (these are far from exhaustive). 

\subsection{Global degeneracies}

\paragraph{Matrix-sandwich symmetry in deep linear networks.}
For a DLN with composite weight $W=W_M\cdots W_1$ one can insert any
invertible matrix $O$ between any neighboring layers, e.g.,
$
   W=(W_M O)(O^{-1}W_{M-1})W_{M-2}\!\dotsm W_1,
$
without changing the implemented function.
This produces a $\,\mathrm{GL}(H)$–orbit of equivalent parameters.

\paragraph{ReLU scaling symmetries.}
Because $\operatorname{ReLU}(ax)=a\,\operatorname{ReLU}(x)$ for all $a>0$,
scaling pre–activation weights by $a$ and post–activation weights by $a^{-1}$
leaves the network invariant.
Such \emph{positively scale–invariant} (PSI) directions invalidate naïve
flatness measures
\citep{yi2019psi,pmlr-v119-tsuzuku20a}.

\paragraph{Permutation and sign symmetries.}
Exchanging hidden units (or simultaneously flipping signs of incoming and outgoing
weights) are examples of discrete changes that leave network outputs unchanged \citep{carroll2021phase}. 

\paragraph{Batch- and layer-normalization scaling symmetries.} BN and LN outputs do not change when their inputs pass through the same affine map~\citep{van2017l2}.

\subsection{Local and data-dependent degeneracies}

\paragraph{Low-rank DLNs.}
When the end-to-end matrix $W$ is rank-deficient, any transformation restricted to
the null space can be absorbed by the factors $W_\ell$.

\paragraph{Elimination singularities.} If incoming weights to a given layer are zero, the associated outgoing weights are free to take any value. The reverse also holds: if outgoing weights are zero, this frees incoming weights to take any value. Residual or skip connections can help to bypass these degeneracies~\citep{orhan2017skip}.

\paragraph{Dead (inactive) ReLUs.}
If a bias is large enough that a ReLU never activates or if the data distribution is such that the pretraining activations never exceed the bias, then the outgoing weights
become free parameters; they can be set to arbitrary values without affecting loss because they are always multiplied by a zero activation. Incoming weights also become free parameters (up until the point that they change the preactivation distribution enough to activate the ReLU).  

\paragraph{Always-active ReLUs.}
Conversely, ReLUs that are \emph{always on} behave linearly.  In this regime, the incoming and outgoing weight matrices act as a DLN with the associated matrix-sandwich and low-rank degeneracies discussed above.

\paragraph{Overlap singularities.} If two neurons share the same incoming weights, then the outgoing weights become non-identifiable: in this regime, only the sum matters to the model's functional behavior~\citep{orhan2017skip}.

\clearpage

\section{The Geometry of Deep Linear Networks} \label{sec:geometry-of-dlns}

In this section we give the geometric background underlying the parameter-generation method
described in \cref{sec:generating-parameters}, stating in full the results we use to obtain
ground-truth local learning coefficients and giving the integer optimization problem used for
generic parameters. The algorithms referenced in the main text are collected in
\cref{sec:dln-orbit-algorithms}.

\subsection{Background} \label{sec:background-geometry-of-dlns}

The geometry of the set of optimal parameters of a deep linear network for the
task described in \cref{sec:deep-linear-network-task} is non-trivial. Here, we summarize 
the results necessary to explain our method for generating different types of optimal parameters and 
compute ground-truth values for the local learning coefficient. For a detailed study of the 
geometry of deep linear networks see 
\citet{aoyagiConsiderationLearningEfficiency2024,lehalleurGeometryFibersMultiplication2024}. 

Fix a number of layers $N$ and layer widths $\ve d = (d_0, \ldots, d_N)$ and consider the 
deep linear network $F(-; \ve W) : \R^{d_0} \to \R^{d_N}$ (\cref{def:deep-linear-network})
parametrized by matrices $\ve W = (W_1, \ldots, W_N)$, where $W_i$ is a $d_i \times d_{i-1}$
matrix. For a $d_N \times d_0$ matrix $B$, we define the function $G: \R^{d_0} \to \R^{d_N}$ by $G(x) = Bx$, 
and consider the Bayesian learning regression task described in \cref{sec:deep-linear-network-task}. 
That is, we fix an input distribution $q(x)$ on $\R^{d_0}$ and consider the statistical model 
defined in \eqref{eqn:model-regression-task}. Let $\mathcal{W}_0$ be the set of
parameters $\ve W$ which satisfy $F(x; \ve W) = G(x)$. This can be written as 
\[
    \mathcal{W}_0 = \left\{\ve W \mid L(\ve W) = 0\right\}
\]
where $L(\ve W)$ is defined in \eqref{eqn:expected-negative-log-likelihood-regression-task}. 

The first result concerns the minimum value of the local learning coefficient
$\lambda(\ve W)$ for $\ve W \in \mathcal{W}_0$. Recall from \cref{sec:slt-background}
that $\lambda(\ve W)$ can be understood as the rate of local volume scaling in parameter space 
in a neighborhood of $\ve W$; see \citet[Definition 1]{lauLocalLearningCoefficient2024}
for a precise definition. The minimum $\lambda(\ve W)$ over $\mathcal{W}_0$
\[
    \lambda = \inf _{\ve W \in \mathcal{W}_0} \lambda (\ve W)
\]
is the \emph{(global) learning coefficient} and is a central quantity in singular 
learning theory \citep{watanabeAlgebraicGeometryStatistical2009}. For regression using 
deep linear networks the 
global learning coefficient is known due to \citet{aoyagiConsiderationLearningEfficiency2024}. 

\begin{theorem}[{\citealp[Theorem 1]{aoyagiConsiderationLearningEfficiency2024}}]
    \label{thrm:aoyagi-dln-apdx}
    Consider a $N$-layer DLN with layer sizes 
    $d_0, \ldots, d_N$, and the regression task described in \cref{sec:deep-linear-network-task}.
    Let $r = \rank(B)$ be the rank of the matrix defining the true function, and define
    $\Delta _i = d_i - r$. There exists 
    a set of indices $\Sigma \subseteq \{0, 1, \ldots, N\}$ which 
    satisfies:
    \begin{enumerate}
        \item $ \max \left\{\Delta _\sigma \mid \sigma \in \Sigma\right\} < \min \left\{ \Delta _{\overline{\sigma}} \mid \overline{\sigma} \in \Sigma ^c\right\}$
        \item $ \sum _{\sigma \in \Sigma} \Delta _{\sigma} \geq \ell \cdot \max \left\{ \Delta _\sigma \mid \sigma \in \Sigma \right\} $
        \item $ \sum _{\sigma \in \Sigma} \Delta _{\sigma} < \ell \cdot \max \left\{ \Delta _{\overline{\sigma}} \mid \overline{\sigma} \in \Sigma ^c \right\} $
    \end{enumerate}
    where $\ell = |\Sigma | - 1$ and 
    $\Sigma ^c = \{0, 1, \ldots, N\} \setminus \Sigma$. 
    The global learning coefficient $\lambda$ is:
    \[
        \lambda = \tfrac{1}{2}(r(d_0 + d_N) - r^2) 
        + \tfrac{1}{4\ell} a (\ell - a) 
        - \tfrac{1}{4\ell} (\ell - 1) \left(\sum _{\sigma \in \Sigma} \Delta_\sigma \right) ^2 + \tfrac{1}{2} \sum _{(\sigma, \sigma') \in \Sigma ^{(2)}} \Delta _{\sigma} \Delta _{\sigma '}
    \]
    where $a = \sum _{\sigma \in \Sigma} \Delta _\sigma - \ell \left(\left\lceil 
    \tfrac{1}{\ell}\sum _{\sigma \in \Sigma} \Delta _\sigma \right\rceil - 1 \right)$, $x \mapsto \lceil x \rceil$ is the ceiling function, and 
    $\Sigma ^{(2)} = \{(\sigma, \sigma ' ) \in \Sigma\times \Sigma \mid \sigma < \sigma'\}$ is the set of all 
    2-combinations of $\Sigma$. 
\end{theorem}

\cref{thrm:aoyagi-dln-apdx} provides a ground-truth value for the local learning coefficient for 
the parameters which attain the global learning coefficient, but does not tell us about the local 
learning coefficient at other parameters, or tell us which parameters attain the 
global minimum. This is addressed using the ideas of \citet{lehalleurGeometryFibersMultiplication2024}.

Given $N$ and $\ve d = (d_0, \ldots, d_N)$ above, we denote by $\Rep_{\ve d}$ the vector space 
of all tuples of matrices $\ve W = (W_1, \ldots, W_N)$
where $W_i$ is a $d_{i} \times d_{i-1}$ matrix with real entries. Note that $\Rep_{\ve d}$
is the set of all deep linear network parameters. In 
\citet{lehalleurGeometryFibersMultiplication2024}, the geometry of $\mathcal{W}_0$ is 
related to a group action on $\Rep_{\ve d}$. 
Let $\GL_{d_i}$ denote the group of $d_i\times d_i$ invertible matrices and 
consider the product group $G = \prod _{i=0} ^N \GL _{d_i}$. This acts on 
$\Rep _{\ve d}$ via change of basis: given $g = (P_0, \ldots, P_N) \in G$
and $\ve W \in \Rep_d$, the action of $g$ on $\ve W$ is
\[
    g \cdot \ve W = (P_1 W_1 P_0^{-1}, P_2 W_2 P_1 ^{-1}, \ldots, P_N W_N P_{N-1} ^{-1}). 
\]
Recall that the \emph{G-orbit} of an element $\ve W \in \Rep _{\ve d}$ is the set 
\[
    \{\ve W' \in \Rep _{\ve d} \mid 
    \text{there exists}~g\in G~\text{such that}~\ve W' = g \cdot \ve W
    \} .
\]

\citet{lehalleurGeometryFibersMultiplication2024} exploit parametrizations of 
$G$-orbits in terms of two types of combinatorial objects:
Kostant partitions and rank patterns.
\begin{definition} \label{def:kostant-partition-rank-pattern}
    Let $\ve{d} = (d_0, \ldots, d_N)$ be positive integers. 
    \begin{enumerate}
        \item A \emph{Kostant partition} is an $(N+1)\times (N+1)$ upper triangular matrix 
        $M = (m_{ij}) _{0 \leq i, j \leq N}$ with non-negative integer entries that 
        satisfy
        \[  
            d_k = \sum _{i \leq k \leq j} m_{ij}
        \]
        for all $k = 0, \ldots, N$. 
        \item A \emph{rank pattern} is 
        an $(N+1)\times (N+1)$ upper triangular matrix $R = (r_{ij}) _{0 \leq i, j \leq N}$
        with non-negative integer entries which satisfy
        \[
            d_k = r_{kk}
        \]
        for all $k=0, \ldots, N$ and 
        \begin{equation} \label{eqn:rank-pattern-coherence-constraint}
            r_{ij} - r_{i, j+1} - r_{i-1,j} + r_{i-1,j+1} \geq 0
        \end{equation}
        for all $0 \leq i \leq j \leq N$, where in \eqref{eqn:rank-pattern-coherence-constraint}
        we define $r_{ij} \defeq 0$ if $i < 0$ or $j > N$. 
    
    \end{enumerate}
    The set of all Kostant partitions is denoted by $\mathcal{M} ^+ _{\ve{d}}$ and the set of 
    all rank patterns is denoted by $\mathcal{R}^\text{orb} _{\ve d}$.
\end{definition}

The set of $G$-orbits, rank patterns and Kostant partitions are in bijection. It is convenient 
to represent $G$-orbits by elements of $\Rep_{\ve d}$ a particular form; a \emph{partial permutation matrix} is 
a matrix where all entries are either 0 or 1 and each row and column contains at most one 1. Each $G$-orbit can 
be represented by a vector of partial permutation matrices. 

We define two functions $\rho : \Rep _{\ve d} \to \mathcal{R}^{\textnormal{orb}}$ and 
$\mu : \mathcal{R}^\textnormal{orb}_{\ve d} \to \mathcal{M} ^+ _{\ve{d}}$ as follows. For 
$\rho$ let 
\[
    r_{ij}(\ve W) = \rank (W_j W_{j-1} \cdots W_i)
\]
for $i < j$ and $r_{ii}(\ve W) = d_{ii}$, and define $\rho(\ve W) = (r_{ij}(\ve W))$. 
For $\mu$ let
\[
    m_{ij}(R) = r_{ij} - r_{i, j+1} - r_{i-1,j} + r_{i-1,j+1}   
\]
and define $\mu(R) = (m_{ij}(R))$.

\begin{lemma} \label{lemma:orbit-bijection}
    We have:
    \begin{enumerate}
        \item The function $\rho : \Rep _{\ve d} \to \mathcal{R}^{\textnormal{orb}}$
        is surjective and has the property that 
        $\ve W$ and $\ve W'$ are in the same $G$-orbit if and only if $\rho(\ve W) = \rho(\ve W')$. 
        That is, $\rho$ induces a bijection between $\mathcal{R}^{\textnormal{orb}}$ and
        $G$-orbits of $\Rep _{\ve d}$. 
        \item The function $\mu : \mathcal{R}^\textnormal{orb}_{\ve d} \to \mathcal{M} ^+ _{\ve{d}}$
        is a bijection between rank patterns and Kostant partitions. 
    \end{enumerate}
    \begin{proof}
        See \citet{lehalleurGeometryFibersMultiplication2024}.
    \end{proof}
\end{lemma}

Certain aspects of the geometric structure of $\mathcal{W}_0$ --- specifically the relationship between 
irreducible components of $\mathcal{W}_0$ --- can be understood via a partial order on the set 
$\mathcal{R}^{\textnormal{orb}}$ of rank patterns. For rank patterns $R$ and $S$ we define 
\begin{equation} \label{eqn:rank-pattern-partial-order}
    R \leq S \qquad \text{if and only if}\qquad r_{ij}\leq s_{ij} ~\text{for all}~ i \leq j .
\end{equation}
We say a rank pattern $S$ is \emph{maximal} (resp. \emph{minimal}) 
if it is maximal (resp. minimal) with respect to the partial order. 

\begin{theorem}  \label{thrm:special-true-parameter-characterisation}
    Suppose $d_i = d$ for all $i=0, \ldots, N$ and let $r = \rank (B)$ be the rank of the 
    true function. Among the rank patterns corresponding to orbits intersecting with $\mathcal{W}_0$, 
    the rank pattern
    \[
        R = \begin{pmatrix}
            d & r & r & \cdots & r & r \\
              & d & r & \cdots & r & r \\
              &   & d & \cdots & r & r \\
              &   &   & \ddots &   & \vdots \\
              &   &   &        & d & r \\
              &   &   &        &   & d \\
        \end{pmatrix}
    \]
    is minimal. The corresponding $G$-orbit is represented by an element $\ve W = (W_1, \ldots, W_N)$ 
    where each $W_i$ is a partial permutation matrix where the first $r$ rows have 
    a 1 on the diagonal and all remaining entries are 0. All elements of $\mathcal{W}_0$ in this 
    $G$-orbit have local learning coefficient equal to the global learning coefficient. 
    \begin{proof}
        See \citet[Theorem 3.8, Corollary 4.4]{lehalleurGeometryFibersMultiplication2024}.
    \end{proof}
\end{theorem}

\cref{thrm:special-true-parameter-characterisation} allows us to generate elements of 
$\mathcal{W}_0$ whose local learning coefficient can be computed via the formula given in 
\cref{thrm:aoyagi-dln-apdx}.
Additionally, the parameters described in 
\cref{thrm:special-true-parameter-characterisation} have multiplicity
(see \cref{sec:slt-background}) greater than one, and sit at the intersection of many irreducible 
components of $\mathcal{W}_0$. Hence, these parameters should be understood as the ``most degenerate'' 
parameters in $\mathcal{W}_0$. 

\begin{theorem} \label{thrm:llc-from-orbits}
    Suppose $d_i = d$ for all $i=0, \ldots, N$ and let $r = \rank (B)$ be the rank of the 
    true function. Let $\ve W \in \mathcal{W}_0$ 
    and $R$ be the rank pattern corresponding to its orbit.  
    Consider the set
    \[
    \mathcal{S} = \{S \in \mathcal{R}^{\textnormal{orb}} \mid S \geq R^* ~\text{and}~
    S~\text{is maximal}\}
    \]
    of all maximal rank patterns greater than $R$. Then, the 
    local learning coefficient of $\ve W$ is 
    \[
        \lambda(\ve W) = \frac{1}{2} \left(
            \min _{S \in \mathcal{S}} c(\mu^{-1}(S)) + r (2d - r)
        \right)
    \]
    where $\mu : \mathcal{R}^\textnormal{orb} \to \mathcal{M} ^+ _{\ve{d}}$ is 
    the bijection in \cref{lemma:orbit-bijection} and 
    \[
        c(M) = \sum _{1 \leq i\leq u\leq j \leq v \leq N} m_{i-1,j-1} m_{uv}
    \]
    for a Kostant partition $M$.
\end{theorem}

\cref{thrm:llc-from-orbits} allows us to compute the local learning coefficient of 
an arbitrary point on $\mathcal{W}_0$ provided we know the corresponding rank pattern and 
can find the set $\mathcal{S}$. 

\begin{remark}
    \citet[Theorem 8.6]{lehalleurGeometryFibersMultiplication2024} observe that learning 
    coefficients in deep linear networks are equal to half of the local codimension, while in 
    general this is only an upper bound. This means that deep linear networks do not capture the 
    full extent of degeneracy that is possible in learning machines. This must be taken into 
    account when extrapolating experimental results in deep linear networks to other settings.
\end{remark}

\subsection{Generating special and generic optimal parameters} \label{sec:generating-special-and-generic-params}

We consider two distinct methods for generating elements of $\mathcal{W}_0$.

\begin{description}
    \item[Special parameters:] We consider a method for generating only the most 
        degenerate points in $\mathcal{W}_0$, which we refer to as \emph{special parameters.} 
        These points have local learning coefficient equal to the global learning coefficient 
        and so it can be computed using \cref{thrm:aoyagi-dln-apdx}. 
    \item[Generic parameters:] We consider a method for generating an arbitrary point in 
        $\mathcal{W}_0$, which we refer to as a \emph{generic parameter}. This is done in 
        such a way so that it is possible to compute the local learning coefficient using
        \cref{thrm:llc-from-orbits}.
        In theory this method can produce any point on $\mathcal{W}_0$, including the special 
        parameters discussed above, but on average parameters generated via this method will 
        be less degenerate than the special parameters.
\end{description}

In both the special and generic cases we begin by randomly choosing the number of 
layers $N$ uniformly between $N_\text{min}$ and $N_\text{max}$, and a width $d$ uniformly 
between $d_\text{min}$ and $d_\text{max}$. The values $N_\text{min}$, $N_\text{max}$, 
$d_\text{min}$ and $d_\text{max}$ determine the model class and are given in \cref{tab:dln-architecture-hparams}. We then generate $N$ partial permutation matrices 
$\ve W _\text{ppm} = (W_1, \ldots, W_N)$, where each $W_i$ is $d\times d$. 

\begin{table}[t]
\caption{Deep linear network architecture hyperparameters.}
\label{tab:dln-architecture-hparams}
\begin{center}
\begin{small}
\begin{tabular}{l|ccccccc}
    \toprule
    & Generic & 100K & 1M & 10M & 100M \\
    \midrule
    Minimum number of layers $N_\text{min}$ & 2 & 2 & 2 & 2 & 2\\
    Maximum number of layers $N_\text{max}$ & 5 & 10 & 20 & 20 & 40 \\
    Minimum width $d_\text{min}$ & 5 & 50 & 100 & 500 & 500 \\
    Maximum maximum width $d_\text{max}$ & 50 & 500 & 1000 & 2000 & 3000\\
    \bottomrule
\end{tabular}        
\end{small}    
\end{center}
\end{table}

\begin{itemize}
    \item In the special case we generate a rank $r$ uniformly between $0$ and $d/2$. Each 
    $W_i$ is then taken to be the $d\times d$ matrix where the first $r$ diagonal entries 
    are one and the remainder of the entries are zero. 
    \item In the generic case, 
    we generate each $W_i$ by randomly choosing the number of non-zero entries, forming 
    a diagonal matrix with the specified number of ones on the diagonal, and shuffling the 
    rows and columns. We repeatedly regenerate $(W_1, \ldots, W_N)$ in this manner 
    until the rank of the product $W_N W_{N-1} \cdots W_1$ is non-zero. By generating 
    $W_i$ as a partial permutation matrix we can identify the corresponding rank pattern 
    and Kostant partition, which is necessary to compute the local learning coefficient.
\end{itemize}
To obtain more realistic parameters and introduce more variation into the parameters we 
test, we then randomly generate an element $g \in G$ to apply to $\ve W _\text{ppm}$. 
This allows us to test parameters which are not partial permutation matrices, and 
because $g \cdot \ve W_\text{ppm}$ is in the same $G$-orbit as $\ve W_\text{ppm}$ 
it has the same local learning coefficient and multiplicity. We give the full procedures
for generating the special and generic parameters in 
\cref{alg:generate-special,alg:generate-generic}.

We generate $g = (P_0, \ldots, P_N)$ 
in the same way for both special and generic parameters. To ensure 
$g \cdot \ve W_\text{ppm}$ is numerically stable in the context of both the forward 
pass and sampling, $g$ needs to be generated so that the entries of $g\cdot \ve W_\text{ppm}$
are not too large. This is achieved by generating each $P_i$ by first drawing 
two orthogonal matrices $U_i$ and $V_i$ independently from the Haar measure 
using the algorithm given by \citet[Section 5]{mezzardi2007generateRandomMatrices}.
We then take $P_i = U_i S_i V_i$ where $S_i$ is a diagonal matrix with 
positive diagonal entries $s_0^i, \ldots, s_N^i$ generated independently 
as $s_k^i = \exp(t_k^i)$ where $t_k^i$ is uniformly distributed on $[-a, a]$. 
In all experiments we take $a = 1/2$. This procedure is given in \cref{alg:generate-g}.

The fact that $U_i$ and $V_i$ are generated from the Haar measure and that the 
size of the entries of $S_i$ is controlled by $a$ means $g \cdot \ve W_\text{ppm}$
is numerically well-behaved. 
The variance of each entry of $U_i$, $V_i$ is $1/d$. 
Putting aside the effect of $S_i$, 
when we multiply a layer of $\ve W _\text{ppm}$ by $U_iV_i$, the non-zero entries also 
have variance $1/d$. This is related to the initialization of deep linear networks 
as orthogonal matrices in \citet{saxeExactSolutionsNonlinear2014}, and also matches the 
entrywise variance of other common initialization schemes such as in 
\citet{he2015delving} and \citet{glorot2010understanding}. The matrix $S_i$ 
introduces a dilation to $P_i$; the 
diagonal entries of $S_i$ can be understood as the singular values of $P_i$. 
Setting the 
parameter $a$ sufficiently small is crucial for ensuring the final weights $g\cdot \ve W$ are 
numerically stable. The value $a = 1/2$ used in all experiments was 
set conservatively by trial and error.

\subsection{Computing the local learning coefficient} \label{sec:appendix-computing-llc}

The local learning coefficient is computed differently for special and generic parameters, 
generated via \cref{alg:generate-special} and \cref{alg:generate-generic} respectively.

\paragraph{Computing the local learning coefficient for a special parameter.}
\cref{alg:generate-special}, used to generate the special parameters, is designed 
to produce parameters for which the local learning coefficient is minimal. That is, 
the local learning coefficient coincides with the global learning coefficient 
$\lambda = \inf _{\ve W \in \mathcal{W}_0} \lambda (\ve W)$. This means we can
compute the local learning coefficient using \citet[Theorem 1]{aoyagiConsiderationLearningEfficiency2024}, 
which we restate in \cref{thrm:aoyagi-dln-apdx}. In the notation of \cref{thrm:aoyagi-dln-apdx}, 
the set $\Sigma$ is computed by solving an integer constraint satisfaction problem 
implemented using the library \texttt{OR-Tools} \citep{ortools}. The local learning coefficient can then 
be computed using the formula in \cref{thrm:aoyagi-dln-apdx}.

\paragraph{Computing the local learning coefficient for a generic parameter.}

The local learning coefficient of generic parameters generated using \cref{alg:generate-generic} 
does not necessarily coincide with the global learning coefficient, and so we cannot apply 
\citet[Theorem 1]{aoyagiConsiderationLearningEfficiency2024} in this case. Instead, we 
exploit the relationship between $G$-orbits, combinatorial objects (rank patterns and Kostant partitions), 
and the geometry of $\mathcal{W}_0$ elicited in \citet{lehalleurGeometryFibersMultiplication2024}.

Let $\ve W _\text{ppm}$ be the partial permutation matrices in \cref{alg:generate-generic} used 
to generate the parameter $\ve W$. Because $\ve W$ and $\ve W_\text{ppm}$ are in the same $G$-orbit
they have the same local learning coefficient. First, we compute the rank pattern $R$ corresponding to 
$\ve W _\text{ppm}$ using the bijection in \cref{lemma:orbit-bijection}. 
We then find the set 
\[
    \mathcal{S} = \left\{
        S \mid S~\text{is a rank pattern,}~S \geq R,~\text{and}~S~\text{is maximal}
    \right\}
\]
of maximal rank patterns greater than or equal to $R$ by solving an iteratively constructed 
sequence integer optimization problems. This step is the key bottleneck in scaling experiments 
with generic parameters to larger models. The optimization problems are defined as follows:
\begin{enumerate}
    \item We set up a constrained integer optimization problem with variables 
    $S = (s_{ij})$ which are the entries of a rank pattern and impose the constraints:
    \begin{align}
        s_{0N} = r &\label{eqn:constraint-rank-r} \\
        s_{ii} = d &\qquad ~\text{for all}~ i = 0, \ldots, N  \label{eqn:constraint-width-d} \\
        s_{ij} - s_{i, j+1} - s_{i-1,j} + s_{i-1,j+1} \geq 0 &\qquad ~\text{for all}~ 0 \leq i \leq j \leq N \label{eqn:constraint-rank-pattern}\\
        r_{ij} \leq s_{ij} \leq d &\qquad ~\text{for all}~ 0 \leq i \leq j \leq N  \label{eqn:constraint-greater-than-R}.
    \end{align}
    The set of all matrices satisfying constraints (\ref{eqn:constraint-rank-r}--\ref{eqn:constraint-greater-than-R}) is exactly the set of 
    rank patterns greater than $R$. To find maximal rank patterns maximize the objective function 
    \[
        F(S) = \sum _{i,j} s_{ij} .
    \]
    Any maximal solution $S ^* = s_{ij} ^*$ is a maximal rank pattern greater than $R$. 
    \item Upon finding a solution $S ^* = s_{ij} ^*$, 
    we consider a further constrained optimization problem with variables $S = (s_{ij})$, objective function 
    $F$ and constraints 
    (\ref{eqn:constraint-rank-r}--\ref{eqn:constraint-greater-than-R}) as well as the additional constraint: 
    \begin{equation} \label{eqn:constraint-no-domination}
        s_{ij} > s^*_{ij} \qquad ~\text{for some}~ 0 \leq i \leq j \leq N .
    \end{equation}
    A solution to this new constrained optimization problem is a maximal rank pattern $S^\dagger$ which 
    is greater than $R$. The constraint \eqref{eqn:constraint-no-domination} ensures that 
    $S^\dagger \not\leq S^*$.
    \item This process is repeated, iteratively adding constraints of the form \eqref{eqn:constraint-no-domination}
    as solutions are found, until the constrained optimization problem becomes \emph{infeasible}:
    the set of matrices satisfying all constraints is empty. The set $\mathcal{S}$ of solutions to these 
    iteratively constructed and solved optimization problems is the set of maximal rank patterns greater 
    than $R$.
\end{enumerate}
Given $\mathcal{S}$, we can then compute the local learning coefficient using the formula in \cref{thrm:llc-from-orbits}.
The algorithm for finding maximal rank patterns is given in \cref{alg:find-maximal-rank-patterns}.

This procedure is limited as finding the optimal solution to each constrained optimization problem has 
a worst-case time complexity which is exponential in the number of variables. 
In practice however, there are algorithms which can achieve acceptable 
average time performance for a moderate number of variables. We implement the procedure described above 
using solvers from the library \texttt{OR-Tools} \citep{ortools}. We find that computing the local learning 
coefficient becomes infeasible as $N$ and $d$ are increased due to the growth in the number of 
elements of $\mathcal{S}$ --- and hence the number of optimization problems that must be solved ---
rather than the difficulty of any individual problem. 

\clearpage

\subsection{Algorithms} \label{sec:dln-orbit-algorithms}
\begin{algorithm}[h]
    \caption{Generate Special Parameter} \label{alg:generate-special}
    \begin{algorithmic}
        \STATE {\bfseries Outputs:} $\ve W$ (special parameter in $\mathcal{W}_0$)
        \STATE {\bfseries Hyperparameters:} $N_{\min}, N_{\max}$ (layer bounds), $d_{\min}, d_{\max}$ (width bounds)
        \STATE
        
        \STATE Draw $N$ uniformly from $\{N_{\min}, \ldots, N_{\max}\}$.
        \STATE Draw $d$ uniformly from $\{d_{\min}, \ldots, d_{\max}\}$.
        \STATE Draw $r$ uniformly from $\{0, \ldots, \lfloor d/2 \rfloor\}$.
        \FOR{$i \gets 1:N$}
            \STATE $W_i \gets \operatorname{diag}(\underbrace{1, \ldots, 1}_{r}, \underbrace{0, \ldots, 0}_{d-r})$ \COMMENT{$d \times d$ matrix.}
        \ENDFOR
        \STATE $\ve W_{\mathrm{ppm}} \gets (W_1, \ldots, W_N)$
        \STATE $g \gets \textsc{GenerateGroupElement}(N, d)$ \COMMENT{\Cref{alg:generate-g}.}
        \STATE $\ve W \gets g \cdot \ve W_{\mathrm{ppm}} = (P_1 W_1 P_0^{-1}, \ldots, P_N W_N P_{N-1}^{-1})$
        \STATE \RETURN{$\ve W$}
    \end{algorithmic}
\end{algorithm}

\begin{algorithm}[h]
    \caption{Generate Generic Parameter} \label{alg:generate-generic}
    \begin{algorithmic}
        \STATE {\bfseries Outputs:} $\ve W$ (generic parameter in $\mathcal{W}_0$)
        \STATE {\bfseries Hyperparameters:} $N_{\min}, N_{\max}$ (layer bounds), $d_{\min}, d_{\max}$ (width bounds)
        \STATE
        
        \STATE Draw $N$ uniformly from $\{N_{\min}, \ldots, N_{\max}\}$.
        \STATE Draw $d$ uniformly from $\{d_{\min}, \ldots, d_{\max}\}$.
        \REPEAT
            \FOR{$i \gets 1:N$}
                \STATE Draw $r_i$ uniformly from $\{1, \ldots, d\}$. \COMMENT{Number of non-zero entries.}
                \STATE $D_i \gets \operatorname{diag}(\underbrace{1, \ldots, 1}_{r_i}, \underbrace{0, \ldots, 0}_{d - r_i})$
                \STATE Randomly shuffle the rows and columns of $D_i$ to obtain $W_i$.
            \ENDFOR
        \UNTIL{$\operatorname{rank}(W_N W_{N-1} \cdots W_1) > 0$}
        \STATE $\ve W_{\mathrm{ppm}} \gets (W_1, \ldots, W_N)$
        \STATE $g \gets \textsc{GenerateGroupElement}(N, d)$ \COMMENT{\Cref{alg:generate-g}.}
        \STATE \RETURN{$g \cdot \ve W_{\mathrm{ppm}} = (P_1 W_1 P_0^{-1}, \ldots, P_N W_N P_{N-1}^{-1})$}
    \end{algorithmic}
\end{algorithm}

\begin{algorithm}[h]
    \caption{Generate Group Element} \label{alg:generate-g}
    \begin{algorithmic}
        \STATE {\bfseries Inputs:} $N$ (number of layers), $d$ (width)
        \STATE {\bfseries Outputs:} $g = (P_0, \ldots, P_N) \in G$
        \STATE {\bfseries Hyperparameters:}  $a > 0$ (singular value spread)
        \STATE
        
        \FOR{$i \gets 0:N$}
            \STATE Draw $U_i, V_i \in \R^{d \times d}$ independently from the Haar measure on $O(d)$.
            \FOR{$k \gets 1:d$}
                \STATE Draw $t_k^i$ uniformly from $[-a, a]$.
                \STATE $s_k^i \gets \exp(t_k^i)$
            \ENDFOR
            \STATE $S_i \gets \operatorname{diag}(s_1^i, \ldots, s_d^i)$
            \STATE $P_i \gets U_i S_i V_i$
        \ENDFOR
        \STATE \RETURN{$(P_0, \ldots, P_N)$}
    \end{algorithmic}
\end{algorithm}

\begin{algorithm}[h]
    \caption{Find Maximal Rank Patterns} \label{alg:find-maximal-rank-patterns}
    \begin{algorithmic}
        \STATE {\bfseries Inputs:}  $R = (r_{ij})$ (rank pattern), $N$ (number of layers), $d$ (width), $r$ (rank)
        \STATE {\bfseries Outputs:}  $\mathcal{S}$ (set of maximal rank patterns $\geq R$)
        \STATE
        
        \STATE $\mathcal{S} \gets \emptyset$
        \STATE Define optimization problem $\mathcal{P}$ with variables $S = (s_{ij})_{0 \leq i \leq j \leq N}$:
        \STATE \hspace{\algorithmicindent} \textbf{Objective:} Maximize $F(S) = \sum_{i,j} s_{ij}$
        \STATE \hspace{\algorithmicindent} \textbf{Constraints:}
        \STATE \hspace{\algorithmicindent}\hspace{\algorithmicindent} $s_{0N} = r$
        \STATE \hspace{\algorithmicindent}\hspace{\algorithmicindent} $s_{ii} = d$ for all $i = 0, \ldots, N$
        \STATE \hspace{\algorithmicindent}\hspace{\algorithmicindent} $s_{ij} - s_{i,j+1} - s_{i-1,j} + s_{i-1,j+1} \geq 0$ for all $0 \leq i \leq j \leq N$
        \STATE \hspace{\algorithmicindent}\hspace{\algorithmicindent} $r_{ij} \leq s_{ij} \leq d$ for all $0 \leq i \leq j \leq N$
        \STATE
        \WHILE{$\mathcal{P}$ is feasible}
            \STATE $S^* \gets \textsc{Solve}(\mathcal{P})$
            \STATE $\mathcal{S} \gets \mathcal{S} \cup \{S^*\}$
            \STATE Add constraint to $\mathcal{P}$: $s_{ij} > s^*_{ij}$ for some $0 \leq i \leq j \leq N$
        \ENDWHILE
        \STATE \RETURN{$\mathcal{S}$}
    \end{algorithmic}
\end{algorithm}

\clearpage

\section{Proof of Lemma 2.1} \label{sec:dln-proof}

In this section we prove \cref{thrm:dlns-are-high-degree-polynomials}, 
which shows that the global convergence results for SGLD discussed in \cref{sec:problems-with-global-sampling}
do not apply to the regression problem for deep linear networks described in \cref{sec:deep-linear-network-task}.

Consider an $N$-layer deep linear network $F(-; \ve W) : \R^{d_0} \to \R^{d_N}$ with layer sizes
$d_0, \ldots, d_N$. Recall from \cref{def:deep-linear-network} that this is a family of functions
parametrized by matrices $\ve W = (W_1, \ldots, W_N)$, where $W_l$ is a $d_l \times d_{l-1}$ matrix,
whose end-to-end map is $x \mapsto W x$ with $W = W_N W_{N-1} \cdots W_1$. We consider the regression
task described in \cref{sec:deep-linear-network-task}: the true function is $G(x) = Bx$ for a
$d_N \times d_0$ matrix $B$, the statistical model is given by \eqref{eqn:model-regression-task}, and
the expected and empirical negative log-likelihoods $L(\ve W)$ and $L_n(\ve W)$ are given by
\eqref{eqn:expected-negative-log-likelihood-regression-task} and
\eqref{eqn:empirical-negative-log-likelihood-regression-task} respectively. When sampling from the
tempered posterior distribution \eqref{eqn:tempered-posterior} using a Langevin-based sampler such as
SGLD, $\grad L_n(\ve W)$ is used to compute sampler steps along with the prior and noise terms (see
\cref{sec:problems-with-global-sampling}).

For convenience in the following proof, and in line with the experimental method in
\cref{sec:dln-experiments}, we take the prior to be an isotropic Gaussian distribution on $\ve W$,
\begin{equation}  \label{eqn:dln-prior-isotropic-gaussian}
    \varphi(\ve W) = \left(\frac{\gamma}{2\pi} \right) ^{d/2} \exp\left(
    -\frac{\gamma}{2} \| \ve W - \ve W _\mu \|^2 \right)
\end{equation}
where $\gamma > 0$ and $\ve W _\mu$ is any fixed parameter. By $\| \ve W - \ve W _\mu \|^2$
we mean the sum of the squares of all matrix entries, treating the parameter $\ve W$
in this expression as a vector with $d = \sum _{l=1} ^N d_l d_{l-1}$ entries.

\begin{lemma}[{restating \cref{thrm:dlns-are-high-degree-polynomials}}] \label{thrm:appendix-version-dlns-are-high-degree-polynomials}
    Consider the above situation of an $N$-layer deep linear network learning the described
    regression task. If $q(x)$ is absolutely continuous then with probability one $L_n(\ve W)$ is a degree $2N$
    polynomial in the matrix entries $\ve W = (W_1, \ldots, W_N)$.
    \begin{proof}
        We treat each parameter of the model as a different polynomial variable; that is,
        for each $l=1, \ldots N$ we have $W_l = (w^{l}_{i,j, l})$ where $w_{i, j, l}$ are
        distinct polynomial variables. As before let $W = W_N W_{N-1} \cdots W_1$.
        For a matrix $U$ with polynomial entries we denote
        by $\deg (U)$ the maximum degree of its entries. Hence we have that $\deg (W) = N$
        since each entry of $W$ is a sum of monomials of the form
        $\prod _{l=1} ^N w_{i_l, j_l, l}$. Denote the entries of $W$ by $\mathcal{P}_{ij}(\ve W)$, which
        is a polynomial of degree $N$ in the variables $w_{i, j, l}$. The entries of $B$
        are constants, and we denote them by $b_{ij}$.

        We now consider the
        degree of the likelihood function $L_n(\ve W)$ as a polynomial in the variables
        $w_{i, j, l}$.
        First note that the square of the $i$-th coordinate of $WX_k - BX_k$ is
        \begin{align*}
            (WX_k - BX_k)_i ^2
            &= \left(\sum _{j=1} ^{d_0} \mathcal{P}_{ij}(\ve W) X_k^{(j)} - b_{ij} X_k^{(j)} \right) ^2 \\
            &= \sum _{j=1} ^{d_0} \sum _{j'=1} ^{d_0} X_k ^{(j)} X_{k} ^{(j')} \left(\mathcal{P}_{ij}(\ve W) \mathcal{P}_{ij'}(\ve W)
            - b_{ij} \mathcal{P}_{ij'}(\ve W) - b_{ij'} \mathcal{P}_{ij}(\ve W) + b_{ij} b_{ij'} \right)
        \end{align*}
        where $X_k^{(j)}$ is the $j$-th coordinate of $X_k$. Hence we have
        \begin{align*}
            L_n(\ve W) &= \frac{1}{n}\sum _{k=1} ^n \sum _{i=1} ^{d_N} (WX_k - BX_k)_i ^2 \\
            &= \frac{1}{n}\sum _{k=1} ^n \sum _{i=1} ^{d_N} \sum _{j=1} ^{d_0} \sum _{j'= 1} ^{d_0} X_k ^{(j)} X_{k} ^{(j')} \mathcal{P}_{ij}(\ve W) \mathcal{P}_{ij'}(\ve W) + \left( \text{terms of at most degree } N
            \right)\\
            &= \frac{1}{n} \sum _{i=1} ^{d_N} \sum _{j=1} ^{d_0} \sum _{j'= 1} ^{d_0}  \left( \sum _{k=1} ^n X_k ^{(j)} X_{k} ^{(j')} \right) \mathcal{P}_{ij}(\ve W) \mathcal{P}_{ij'}(\ve W) + \left( \text{terms of at most degree } N
            \right)\\
        \end{align*}
        The polynomial $\mathcal{P}_{ij}(\ve W) \mathcal{P}_{ij'}(\ve W)$ has degree $2N$. The only way $L_n(\ve W)$
        can have degree smaller than $2N$ is if the random coefficients
        $\sum _{k=1} ^n X_k ^{(j)} X_{k} ^{(j')}$ result in cancellation between the terms of
        different $\mathcal{P}_{ij}(\ve W) \mathcal{P}_{ij'}(\ve W)$. We now show that this does not happen with
        probability one. Consider a function $f : \R^{n d_0} \to \R$ given by
        \[
            f(\ve x) = \sum _{(b, b') \in \Lambda} c_{bb'} x_b x_{b'}
        \]
        where $\Lambda$ is any non-empty subset of $\{1, \ldots, n d_0\}^2$ and
        $c_{bb'} \in \R \setminus \{0\}$. The set $f^{-1}(0) \subseteq \R^{n d_0}$ has
        Lebesgue measure zero, and hence has measure zero with respect to the joint distribution
        of the dataset $D_n$ considered as a distribution on $\R^{n d_0}$, since $q(x)$ is
        assumed to be absolutely continuous. It follows that cancellation of the degree
        $2N$ monomials in the above expression for $L_n(\ve W)$ cannot occur with probability
        greater than zero, and thus $\deg (L_n(\ve W)) = 2N$ with probability one.
    \end{proof}
\end{lemma}

Recall from \cref{sec:problems-with-global-sampling}
\citep[and see][Assumption 4]{tehConsistencySGLD2015}
that proofs about the global
convergence of SGLD assume that there exists a function $V : \R^d \to [1, \infty)$
with bounded second derivatives which satisfies
\[
    \|\grad V(\ve W) \|^2 + \| \grad \log \pi (\ve W) \| ^2 \leq C V(\ve W) \qquad \text{ for all } w \in \R^d   .
\]
In the setting of deep linear networks
we have $\grad\log \pi(\ve W) = -n\beta \grad L_n(\ve W)$. Since the second
derivatives of $V$ are bounded and $L_n(\ve W)$ is a degree $2N$ polynomial,
this condition can only hold when $N = 1$. This corresponds precisely to the case
when all critical points of $L_n(w)$ are non-degenerate. Likewise, the global Lipschitz
condition
\[
\| \grad \log \pi(w_1) - \grad \log \pi(w_2) \| \leq \alpha \|w_1 - w_2\| \qquad \text{ for all } w_1, w_2 \in \R^d .
\]
can also only be satisfied when $N=1$.

\begin{remark}
    Consider
    \begin{equation} \label{eqn:posterior-with-true-likelihood}
        \pi(\ve W) \propto \exp(-n\beta L(\ve W)) \varphi(\ve W)
    \end{equation}
    in place of the tempered posterior distribution \eqref{eqn:tempered-posterior}.
    The geometry of $L(\ve W)$ determines
    much of the learning behavior of singular statistical models.
    In SGMCMC algorithms, a stochastic estimate $g(\ve W, U)$ of the
    gradient of the log-posterior could equally be considered an
    estimate of $\grad \log \pi(\ve W)$, where $\pi(\ve W)$ is
    as in \eqref{eqn:posterior-with-true-likelihood}. In the case
    of deep linear networks, a result similar to
    \cref{thrm:appendix-version-dlns-are-high-degree-polynomials} can be shown
    using \eqref{eqn:posterior-with-true-likelihood} in place of the
    usual posterior distribution. In this case we have
    \[
        L(\ve W) = \E \| WX - B X\|^2
    \]
    where $X \sim q(x)$. From the proof of \cref{thrm:appendix-version-dlns-are-high-degree-polynomials}
    we have that
    \begin{align*}
        L(\ve W) &= \E \left[\sum _{i=1} ^{d_N} \sum _{j=1} ^{d_0} \sum _{j'= 1} ^{d_0} X _{j} X _{j'} \mathcal{P}_{ij}(\ve W) \mathcal{P}_{ij'}(\ve W) \right] + \left( \text{terms of at most degree } N \right) \\
        &= \sum _{i=1} ^{d_N} \sum _{j=1} ^{d_0} \sum _{j'= 1} ^{d_0} \E \left[X _{j} X _{j'} \right]  \mathcal{P}_{ij}(\ve W) \mathcal{P}_{ij'}(\ve W) + \left( \text{terms of at most degree } N \right).
    \end{align*}
    where we now write $X_j$ for the $j$-th coordinate of $X$.
    If each coordinate of $X$ is independent and identically distributed (as in the
    experiments in \cref{sec:dln-experiments}) then
    $\E \left[X _{j} ^2 \right] > 0$ and $\E \left[X _{j} X_{j'} \right] \geq 0$ for $j\neq j'$.
    It follows that $L(\ve W)$ has degree $2N$, since all monomial terms with degree $2N$
    in the above expression for $L(\ve W)$ appear with non-negative coefficients, and at least
    some are non-zero.
\end{remark}

\clearpage

\section{Deep Linear Network Experiments} \label{sec:deep-linear-network-appendix}

In this section we give additional details about the deep linear network experiments
The estimation hyperparameters used in the deep linear network experiments are given in \cref{tab:llc-estimation-hyperparameters}.

\begin{table}[h]
\caption{Hyperparameters for local learning coefficient estimation in 
deep linear networks.}
\label{tab:llc-estimation-hyperparameters}
\begin{center}
\begin{small}
\begin{tabular}{l|cccccc}
    \toprule
    Number of sampling steps & $5\times 10^4$ \\
    Number of burn-in steps $B$ & $0.9 T$ \\
    Dataset size $n$ & $10^6$ \\
    Batch size $m$ & $500$ \\
    Localization parameter $\gamma$ & $1$ \\
    Inverse-temperature parameter $\beta$ & $1/\log(n)$\\
    Number of learning problems & $100$ \\
    \bottomrule
\end{tabular}        
\end{small}    
\end{center}
\end{table}.

\subsection{Comparing samplers at special and generic parameters.} \label{sec:generic-vs-special-results}

Here we give additional results from the experiments comparing the performance of samplers at 
special and generic deep linear network parameters. For each sampler, we generate generic and 
special parameters according to the architecture hyperparameters in the `generic' column 
in \cref{tab:dln-architecture-hparams} and use the sampler to estimate the local learning 
coefficient in each case. We use the sampler hyperparameters in \cref{tab:llc-estimation-hyperparameters}. 
For each sampler we used the step size value which resulted in the best accuracy on-average. The step 
size was tuned separately for the special generic and special experiments, though in all cases except 
SGHMC the same step size was used in both cases. 

\begin{figure}[h]
    \centering
    \includegraphics[width=\textwidth]{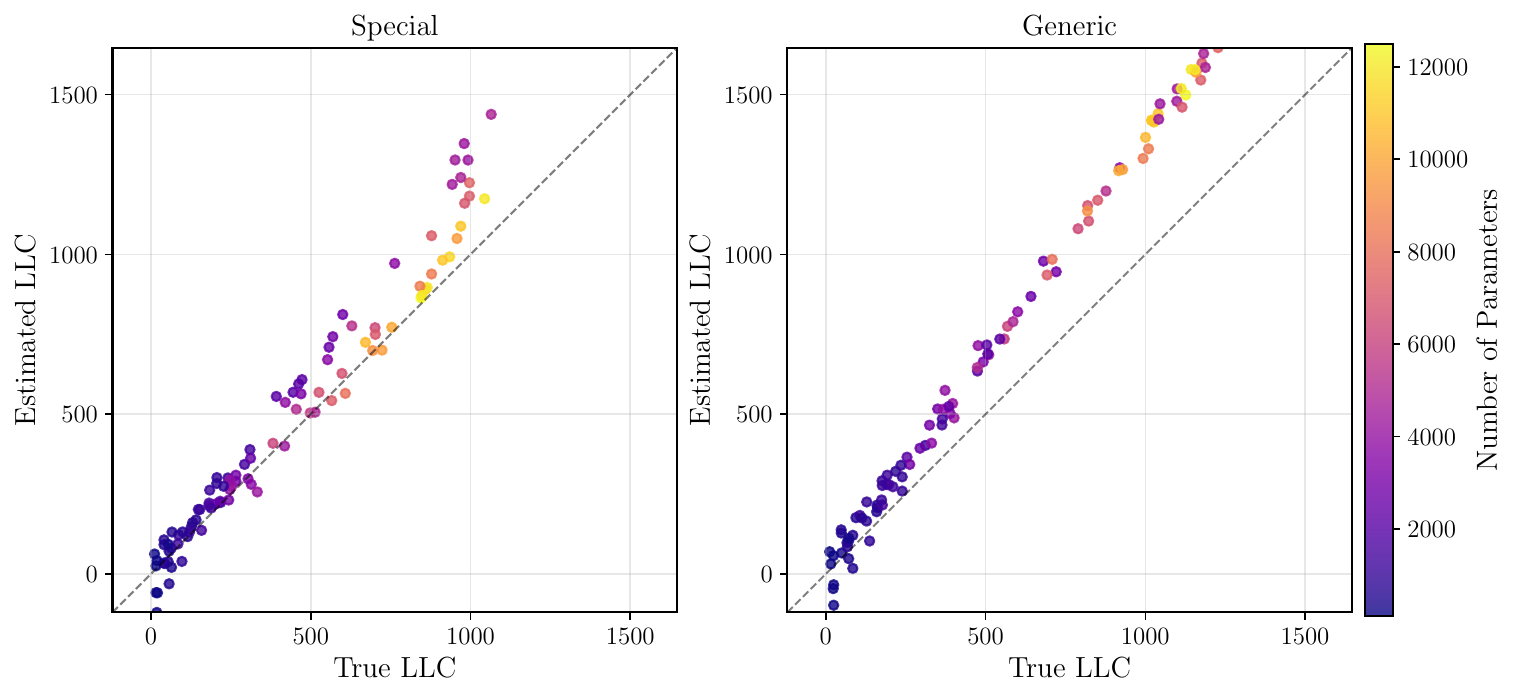}
    \caption{\textbf{AdamSGLD performance on special and generic DLN parameters.} Like RMSPropSGLD, 
    AdamSGLD appears to systematically overestimate the learning coefficient at generic 
    (less degenerate) parameters, and also exhibits some mild overestimation at special 
    (more degenerate) parameters. The variance of estimates is smaller at generic parameters 
    compared to special parameters.}
    \label{fig:adamsgld-generic-vs-special}
\end{figure}

\begin{figure}[h]
    \centering
    \includegraphics[width=\textwidth]{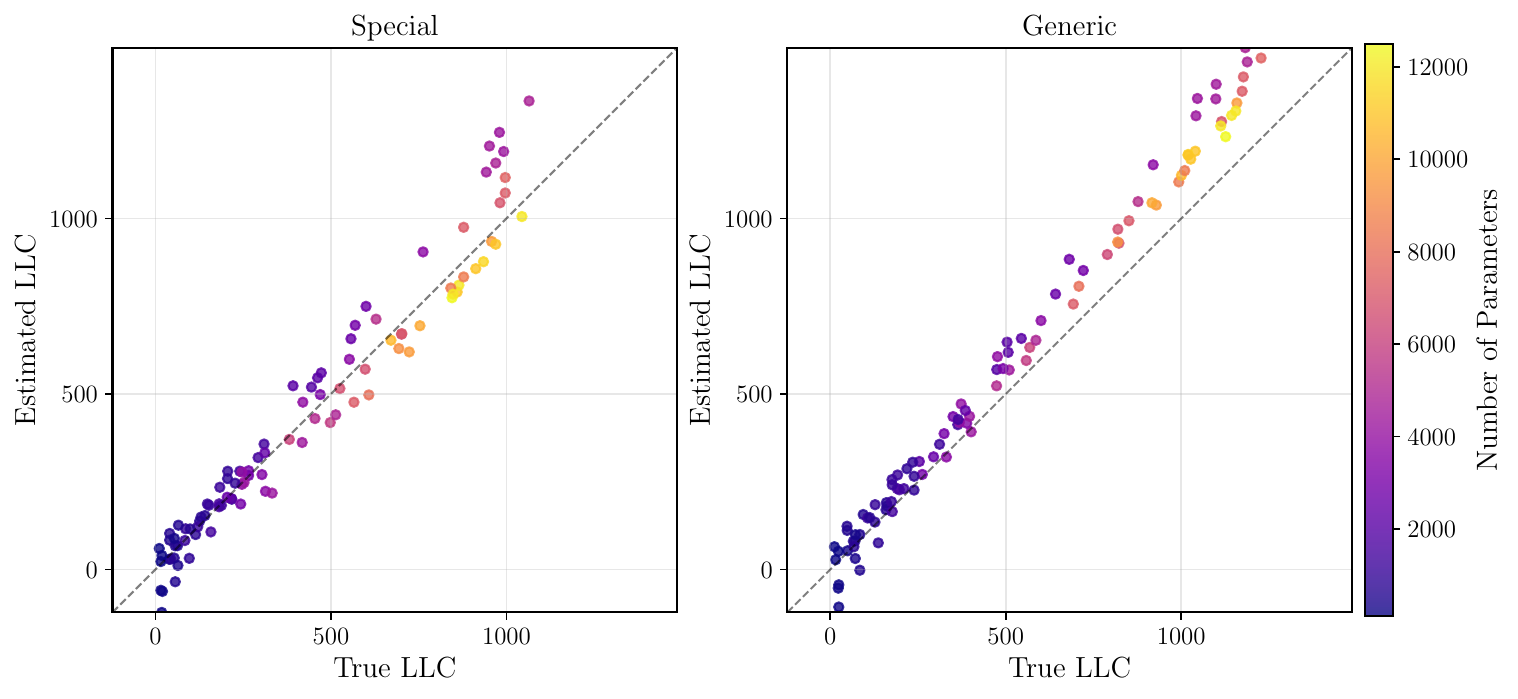}
    \caption{\textbf{RMSPropSGLD performance on special and generic DLN parameters.} Like AdamSGLD, 
    RMSPropSGLD appears to systematically overestimate the learning coefficient at generic 
    (less degenerate) parameters, and also exhibits some mild overestimation at special 
    (more degenerate) parameters. The variance of estimates is smaller at generic parameters 
    compared to special parameters.}
    \label{fig:rmsprop-generic-vs-special}
\end{figure}

\begin{figure}[h]
    \centering
    \includegraphics[width=\textwidth]{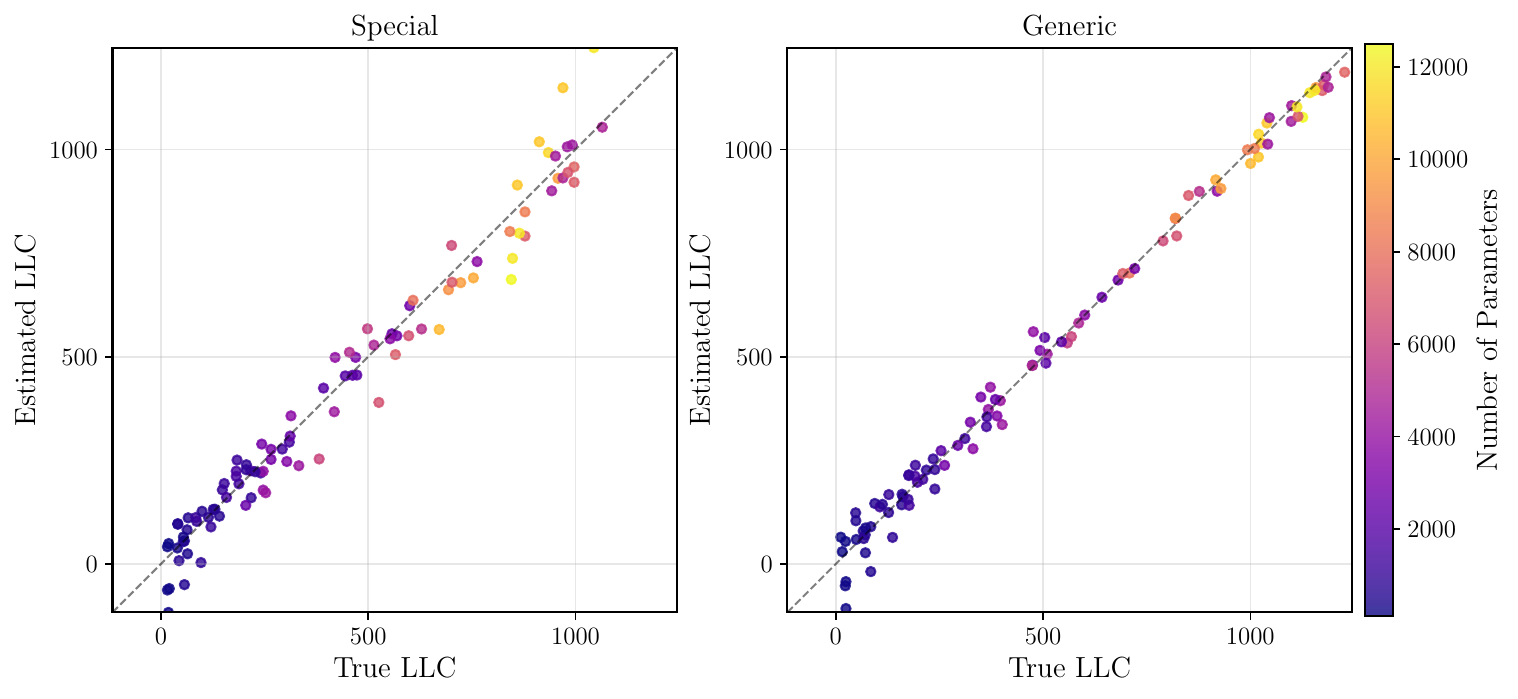}
    \caption{\textbf{SGHMC performance on special and generic DLN parameters.} Like SGNHT, 
    SHGHMC achieves a similar accuracy at both special and generic DLN. Performance appears 
    better at generic parameters, where the estimates are more precise.}
    \label{fig:sghmc-generic-vs-special}
\end{figure}

\begin{figure}[h]
    \centering
    \includegraphics[width=\textwidth]{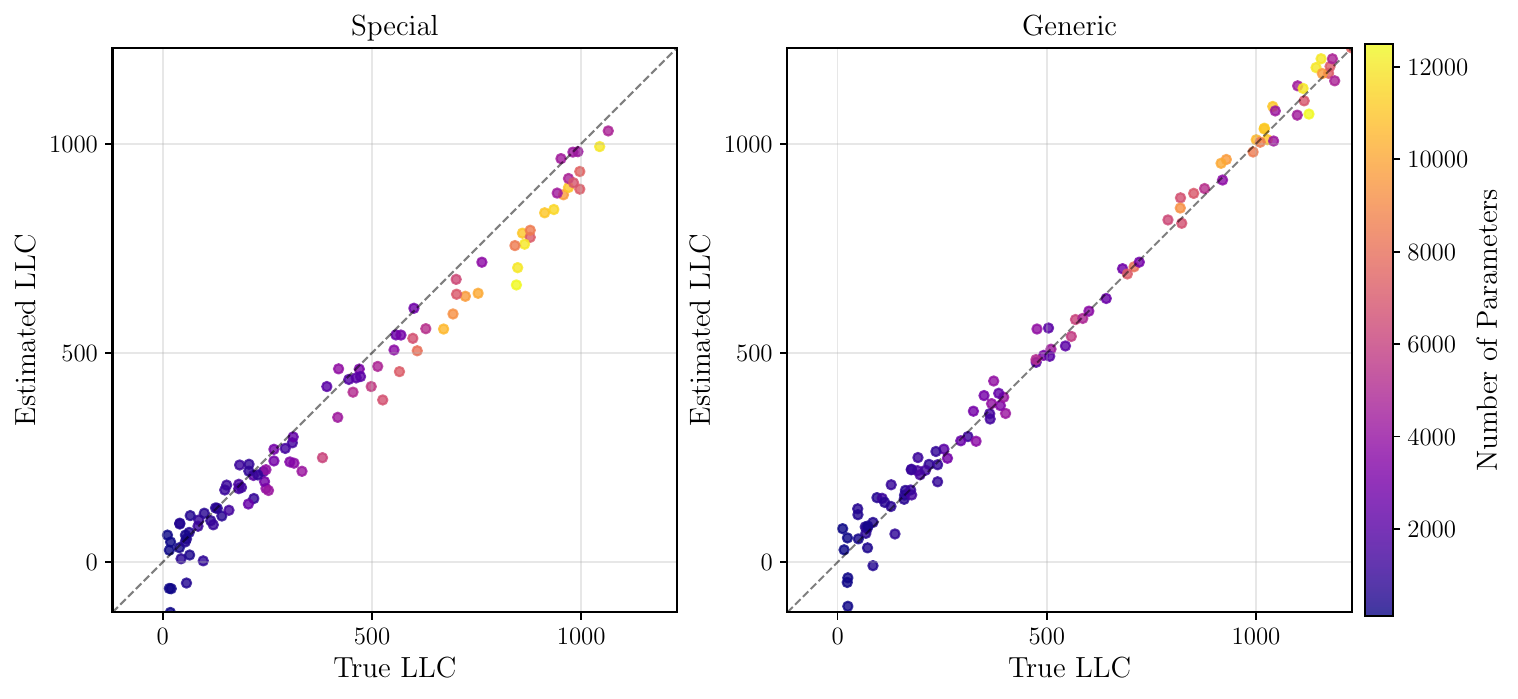}
    \caption{\textbf{SGNHT performance on special and generic DLN parameters.} Like SGHMC, 
    SGNHT achieves a similar accuracy at both special and generic DLN. Performance appears 
    better at generic parameters, where the estimates are more precise.}
    \label{fig:sgnht-generic-vs-special}
\end{figure}

\begin{figure}[h]
    \centering
    \includegraphics[width=\textwidth]{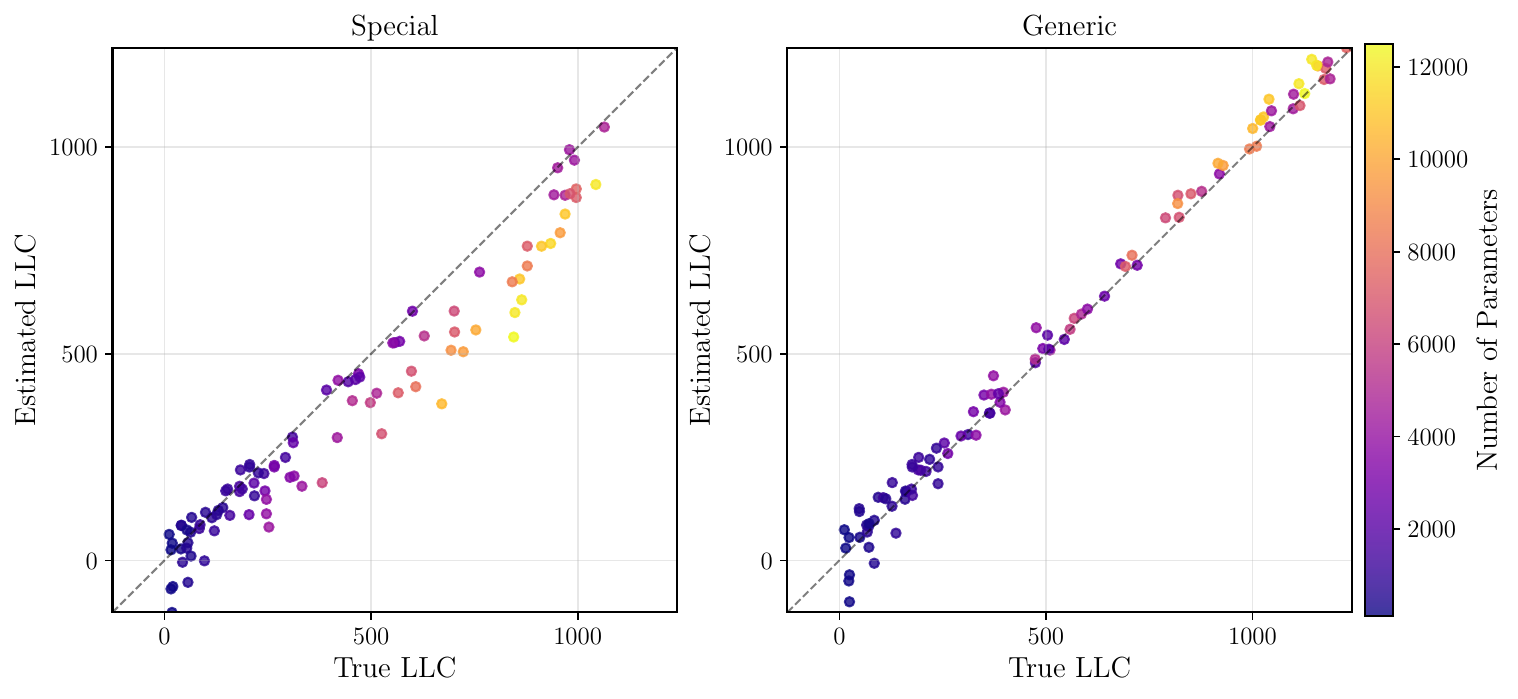}
    \caption{\textbf{SGLD performance on special and generic DLN parameters.} SGLD performs 
    much better at generic parameters, where it achieves similar performance to SGHMC and 
    SGNHT. At special parameters, it underestimates the true local learning coefficient and 
    the estimates have higher variance. Increasing the step size does not 
    significantly improve performance at special parameters, but rather results in more 
    sampling runs encountering numerical issues.}
    \label{fig:sgld-generic-vs-special}
\end{figure}

When plotting the estimated local learning coefficient against the true value we observe that all 
samplers behave differently at special and generic parameters. In all cases, but to varying degrees, 
the LLC estimates at special parameters have higher variance compared to the estimates at generic 
parameters using the same sampler. SGHMC (\cref{fig:sghmc-generic-vs-special}), 
SHNHT (\cref{fig:sgnht-generic-vs-special}) and SGLD (\cref{fig:sgld-generic-vs-special}) all perform better 
at generic (less degenerate) versus special (more degenerate) parameters. In contrast, although their estimates 
have lower variance, both RMSPropSGLD (\cref{fig:rmsprop-generic-vs-special}) 
and AdamSGLD (\cref{fig:adamsgld-generic-vs-special}) both exhibit a systematic overestimation of the LLC 
at generic parameters which does not appear to be present at special parameters. We do not have an explanation 
for this phenomenon. 

Over all five samplers, SGHMC and SGNHT appear to have the best performance across both special and generic parameters. 
We note, however, that the models used in this experiment are smaller than the experiments exclusively 
at special parameters. 
This is because the algorithm used to compute the local learning coefficient at generic parameters 
becomes computationally infeasible for larger models. 
As the model size increases up to 100M parameters --- magnitudes more representative of neural networks 
typically used in practice --- RMSPropSGLD and AdamSGLD both end up outperforming SGHMC and SGNHT 
(see \cref{sec:results,sec:deep-linear-network-appendix-additional-results}) at special parameters. 
Our approach to investigating generic parameters does not scale to higher orders of magnitude, so we 
are unable to conclude whether the same is also true of the generic parameters. 

\clearpage

\subsection{Additional special parameter results} \label{sec:deep-linear-network-appendix-additional-results}

\begin{figure}[h]
    \centering
    \includegraphics[width=\textwidth]{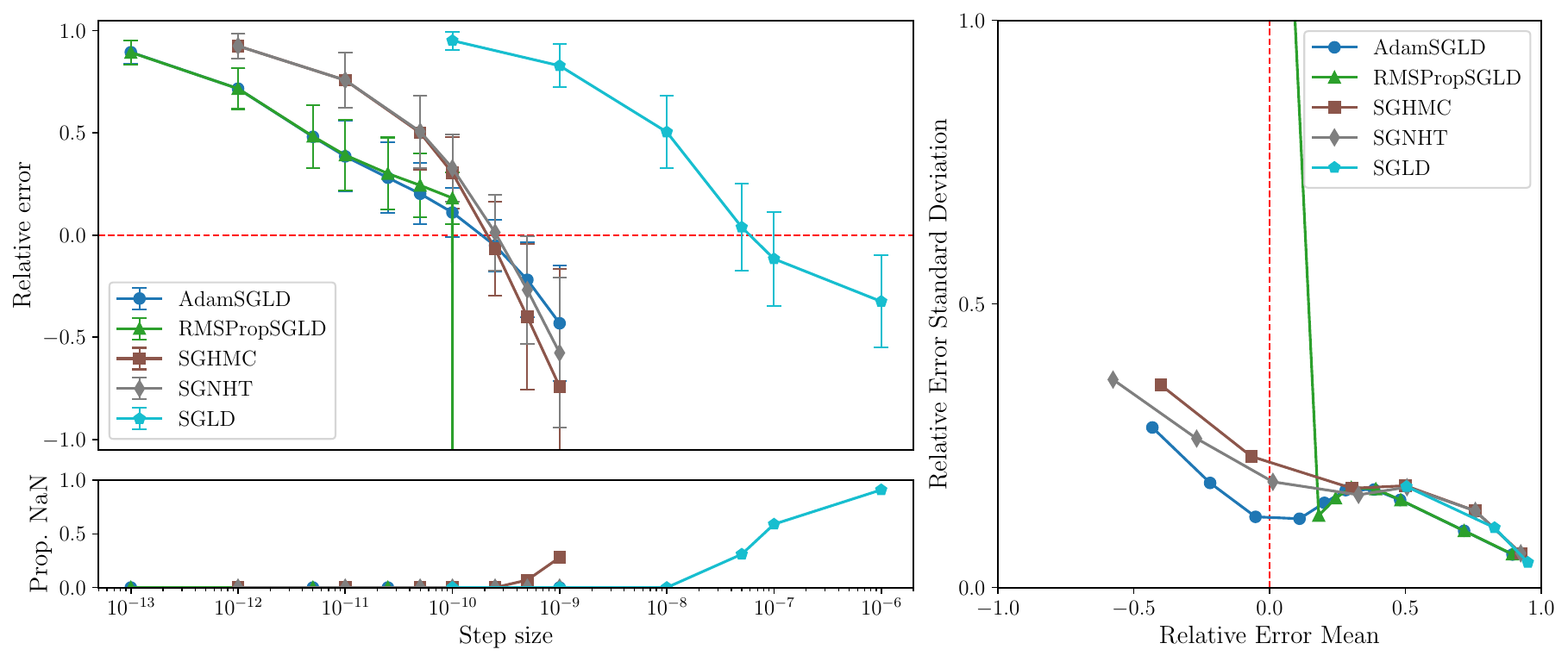}
    \caption{We assess the performance of samplers in estimating the local 
    learning coefficient of 100M parameter deep linear networks. As for the 
    10M parameter models (\cref{fig:relative-error-results-10M}) we see 
    that RMSPropSGLD and AdamSGLD are less sensitive to step size and
    achieve a superior mean-variance trade-off; they remain the best-performing
    samplers at this scale, though their margin over the other samplers is
    narrower than at 10M.}
    \label{fig:relative-error-results-100M}
\end{figure}

\begin{figure}[h]
    \centering
    \includegraphics[width=\textwidth]{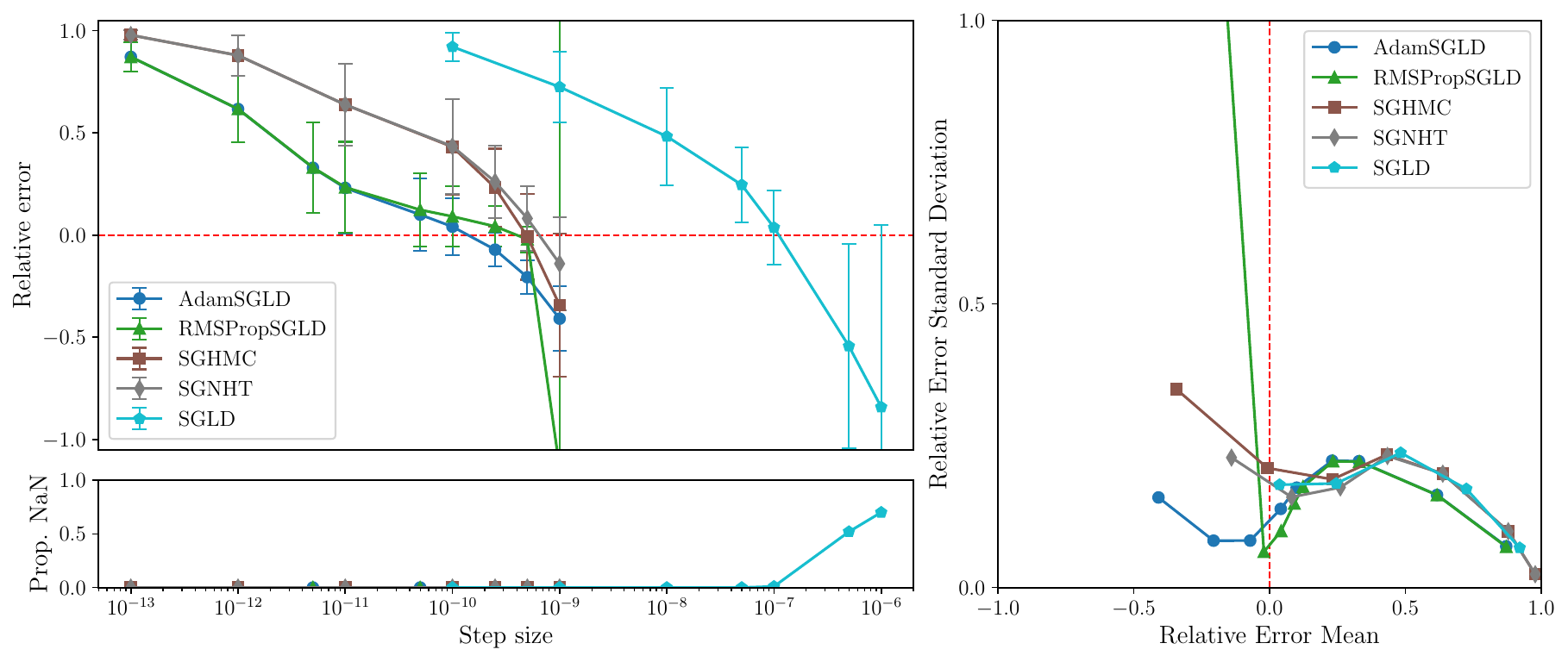}
    \caption{We assess the performance of samplers in estimating the local 
    learning coefficient of 1M parameter deep linear networks. As for the 
    10M parameter models (\cref{fig:relative-error-results-10M}) we see 
    that RMSPropSGLD and AdamSGLD are less sensitive to step size and 
    achieve a superior mean-variance trade-off.}
    \label{fig:relative-error-results-1M}
\end{figure}

\begin{figure}[h]
    \centering
    \includegraphics[width=\textwidth]{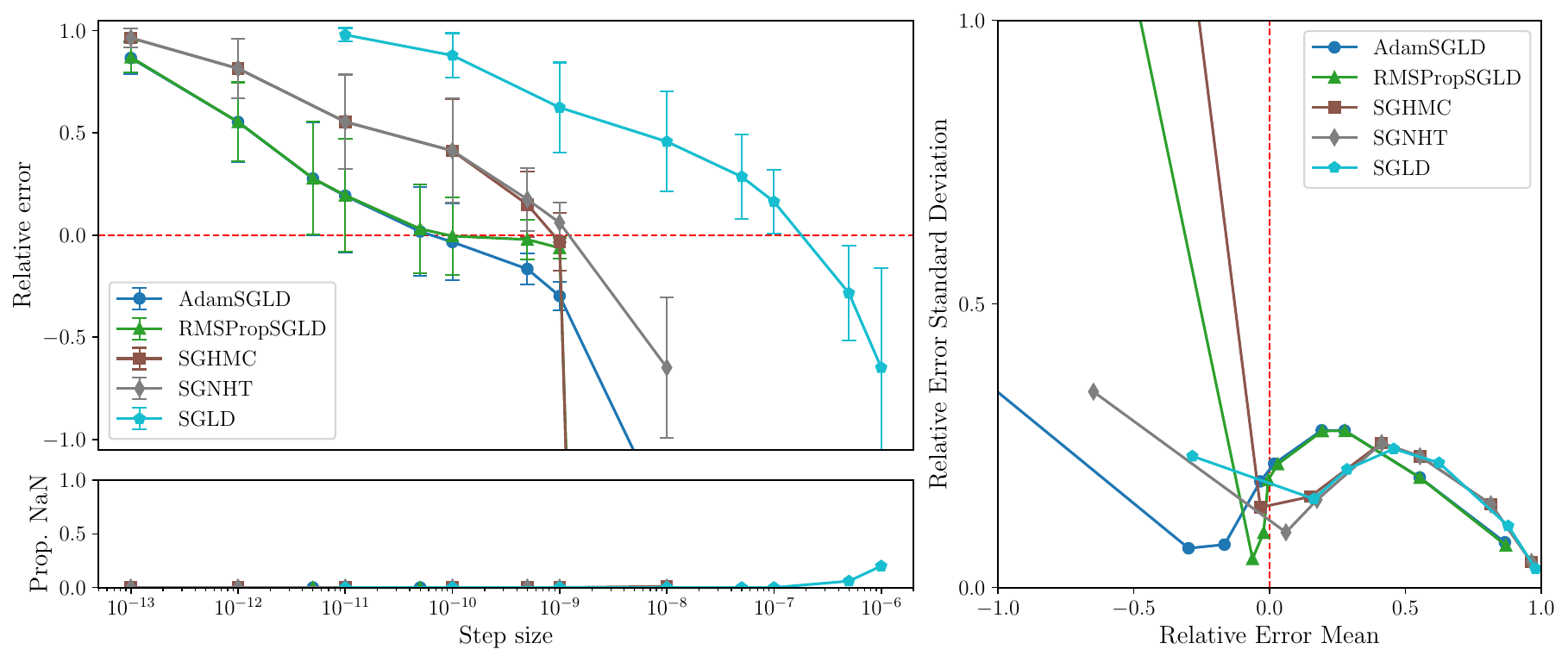}
    \caption{We assess the performance of samplers in estimating the local 
    learning coefficient of 100K parameter deep linear networks. Here RMSPropSGLD
    and AdamSGLD still appear better, though the picture is less clear 
    compared to the 1M, 10M or 100M parameter models. This may suggest that 
    the best choice of sampler may depend on the scale of the sampling problem, 
    though this requires further investigation. }
    \label{fig:relative-error-results-100K}
\end{figure}

Here we give additional results from the experiments with special deep linear network 
parameters described 
in \cref{sec:dln-experiments}. In Figures \ref{fig:relative-error-results-100M},
\ref{fig:relative-error-results-1M} and \ref{fig:relative-error-results-100K}
we present the relative error $(\lambda - \hat \lambda) / \lambda $ in 
the estimated local learning coefficient for the deep linear network 
model classes 100M, 1M and 100K (the results for the 10M model class
are given in \cref{fig:relative-error-results-10M} in the main text).

\begin{figure}[h]
    \centering
    \begin{subfigure}[b]{0.45\textwidth}
        \centering
        \includegraphics[width=\textwidth]{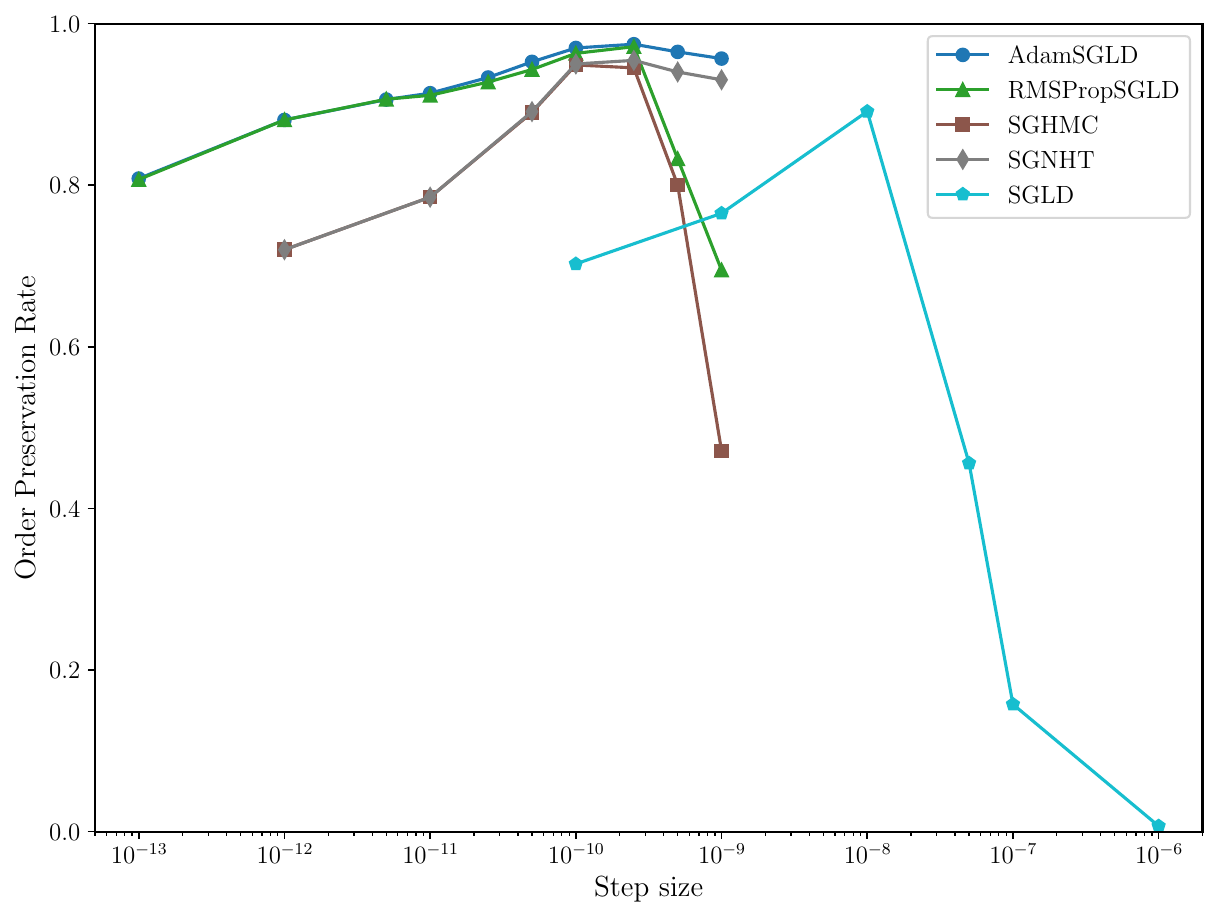}
        \caption{100M}
        \label{fig:order-preservation-100M-subfig}
    \end{subfigure}
    \hfill
    \begin{subfigure}[b]{0.45\textwidth}
        \centering
        \includegraphics[width=\textwidth]{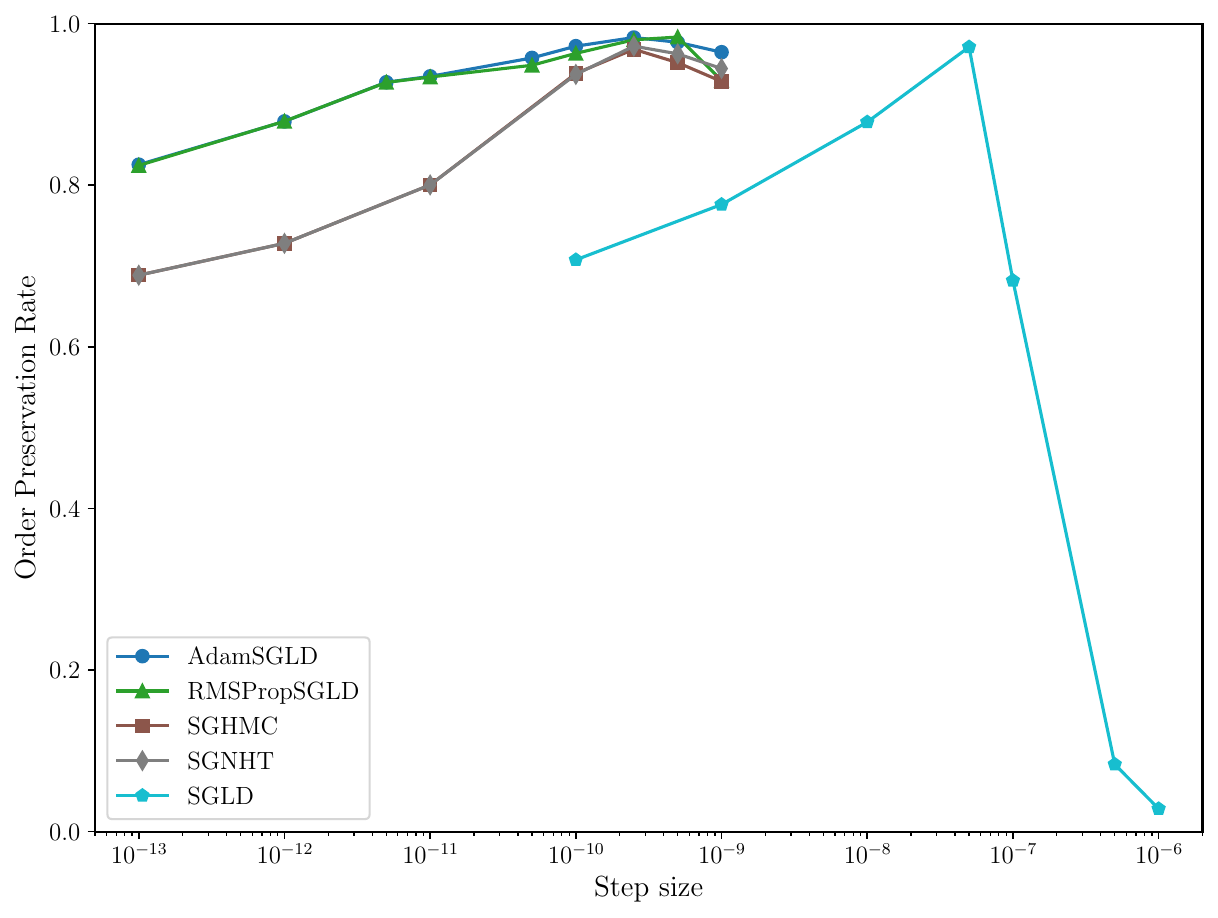}
        \caption{10M}
        \label{fig:order-preservation-10M-subfig}
    \end{subfigure}
    \vskip\baselineskip
    \begin{subfigure}[b]{0.45\textwidth}
        \centering
        \includegraphics[width=\textwidth]{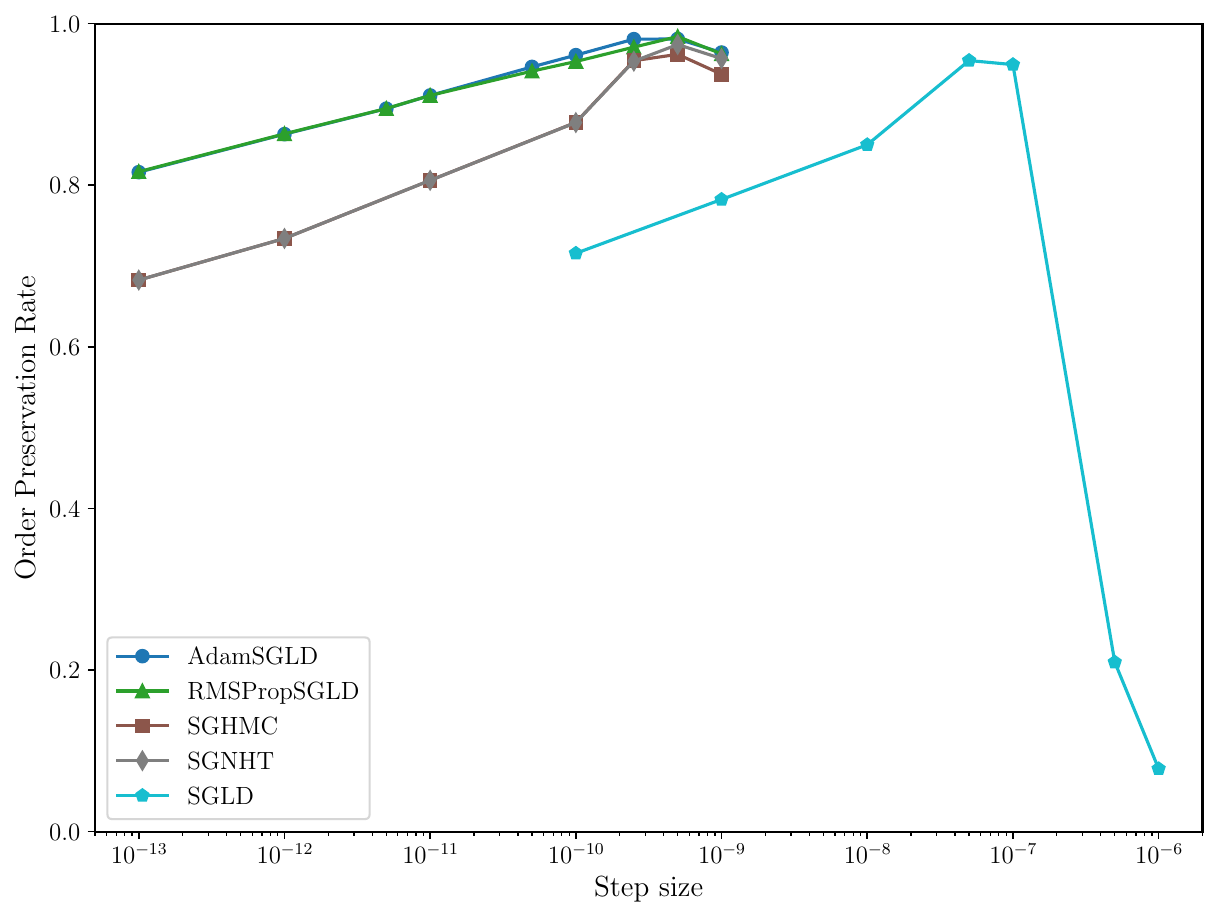}
        \caption{1M}
        \label{fig:order-preservation-1M-subfig}
    \end{subfigure}
    \hfill
    \begin{subfigure}[b]{0.45\textwidth}
        \centering
        \includegraphics[width=\textwidth]{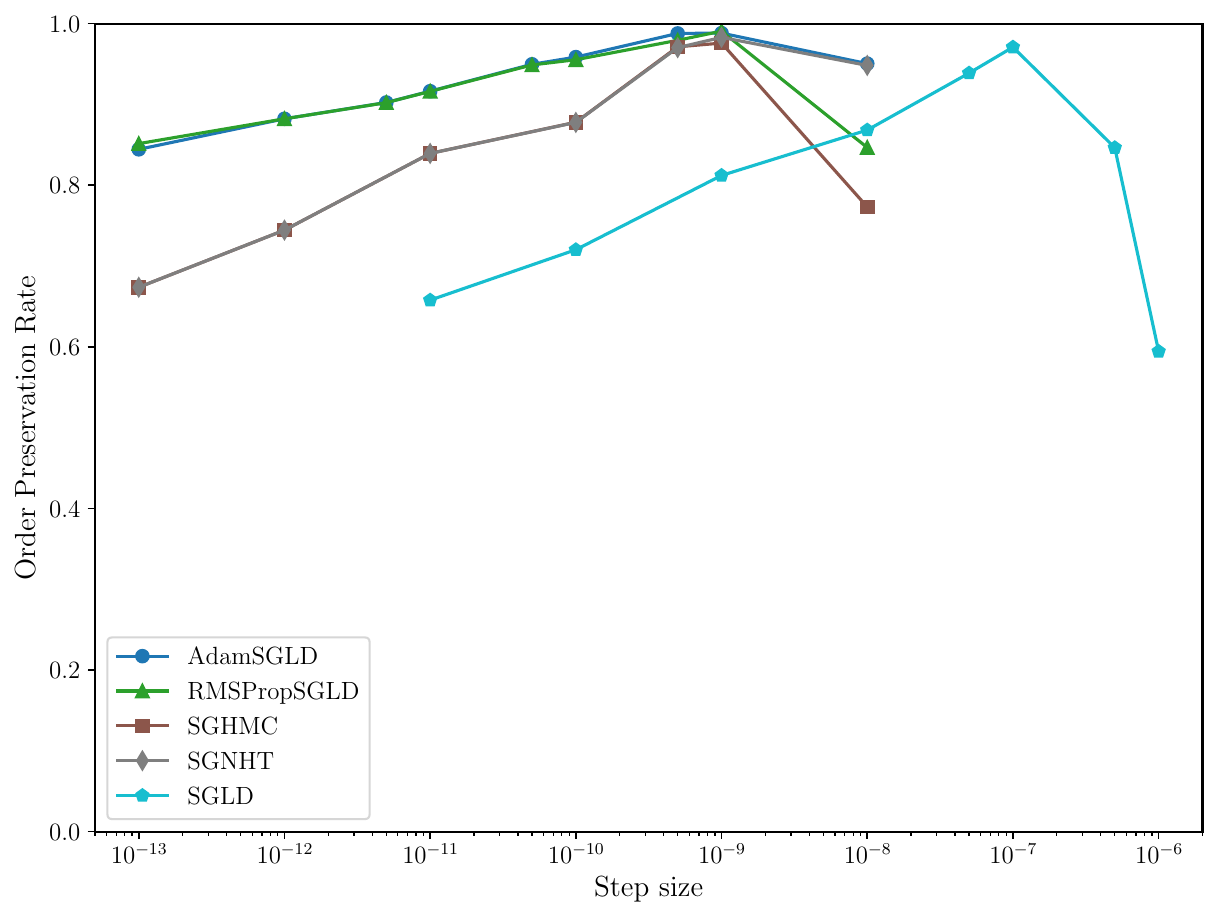}
        \caption{100K}
        \label{fig:order-preservation-100K-subfig}
    \end{subfigure}
    \caption{We assess how well each sampler preserves ordering of true LLCs in the estimated 
    values, reporting the \emph{order preservation rate} as the proportion of pairs of true LLC 
    values $\lambda _1 < \lambda _2$ where the estimates are correctly ordered as $\hat \lambda _1 < \hat \lambda _2$.
    We see that RMSPropSGLD and AdamSGLD are better at preserving order across a wider range of of step 
    sizes than the other samplers for larger models. For RMSPropSGLD and AdamSGLD, good order preservation 
    performances emerges at smaller step sizes compared to accuracy (see 
    Figures \ref{fig:relative-error-results-100M}, \ref{fig:relative-error-results-10M}, \ref{fig:relative-error-results-1M}, 
    \ref{fig:relative-error-results-100K}).}
    \label{fig:order-preservation-rate}
\end{figure}

In \cref{fig:order-preservation-rate} we present the \emph{order preservation
rate} of each sampling algorithm, for each deep linear network model class. This 
assesses how good a sampling algorithm is at preserving the ordering of local 
learning coefficient estimates. These results are discussed in more detail 
in \cref{sec:results} in the main text.

\clearpage

\section{Large Language Model Experiments}  \label{sec:llm-experiments}

In this section we give details of the large language model experiments presented in \cref{fig:llm-rmsprop-vs-sgld}. 
To complement our analysis of deep linear networks, we also examined the performance of sampling algorithms for LLC estimation on a four-layer attention-only transformer trained on the DSIR-filtered Pile \citep{gaoPile800GBDataset2020a,xie2023data}.

\subsection{Model architecture}

We trained a four-layer attention-only transformer with the following specifications:
\begin{itemize}
   \item Number of layers: 4
   \item Hidden dimension ($d_{\text{model}}$): 256
   \item Number of attention heads per layer: 8
   \item Head dimension ($d_{\text{head}}$): 32
   \item Context length: 1024
   \item Activation function: GELU \citep{hendrycks2016gaussian}
   \item Vocabulary size: 5,000
   \item Positional embedding: Learnable (Shortformer-style, \citealt{press2020shortformer})
\end{itemize}

The model was implemented using TransformerLens \citep{nanda2022transformerlens} and trained using AdamW with a learning rate of 0.001 and weight decay of 0.05 for 75,000 steps with a batch size of 32. Training took approximately 1 hour on a TPUv4. 

\subsection{LLC estimation}

We applied both standard SGLD and RMSProp-preconditioned SGLD to estimate the Local Learning Coefficient (LLC) of individual attention heads (``weight-refined LLCs,'' \citealt{wangDifferentiationSpecializationAttention2024}) at various checkpoints during training. As with the deep linear network experiments, we tested both algorithms across a range of step sizes $\epsilon \in \{10^{-4}, 3 \times 10^{-4}, 10^{-3}, 3 \times 10^{-3}\}$. LLC estimation was implemented using \texttt{devinterp}~\citep{devinterpcode}. Each LLC over training time trajectory in~\cref{fig:llm-rmsprop-vs-sgld} took approximately 15 minutes on a TPUv4, for a total of roughly 2 hours. 

\subsection{Results and analysis}

\Cref{fig:llm-rmsprop-vs-sgld} shows LLC estimates for the second head in the third layer. Here, RMSProp-preconditioned SGLD demonstrates several advantages over standard SGLD:

\begin{enumerate}
   \item \textbf{Step size stability}: RMSProp-SGLD produces more consistent LLC-over-time curves across a wider range of step sizes, enabling more reliable parameter estimation.
   
   \item \textbf{Loss trace stability}: The loss traces for RMSProp-SGLD show significantly fewer spikes compared to standard SGLD, resulting in more stable posterior sampling.
   
   \item \textbf{Failure detection}: When the step size becomes too large for stable sampling, RMSProp-SGLD fails catastrophically with NaN values, providing a clear signal that hyperparameters need adjustment. In contrast, standard SGLD might produce plausible but inaccurate results without obvious warning signs.
\end{enumerate}

These results align with our findings in deep linear networks and suggest that the advantages of RMSProp-preconditioned SGLD generalize across model architectures.

\clearpage

\section{Technical Conditions for Singular Learning Theory} \label{sec:slt-technical-conditions-appendix}

In this section we state the technical conditions for singular learning theory (SLT) which are 
required to state \cref{thrm:llc-wbic-estimator}. 
The conditions are discussed in \citet[Appendix A]{lauLocalLearningCoefficient2024} and 
\citet{watanabeAlgebraicGeometryStatistical2009,watanabeMathematicalTheoryBayesian2018}.

\begin{definition}[{\citealp[see][]{watanabeAlgebraicGeometryStatistical2009,watanabe2013WABIC,watanabeMathematicalTheoryBayesian2018}}]
    \label{def:fundamental-slt-conditions}
    Consider a true distribution $q(x)$, model $\{p(x|w)\}_{w\in \mathcal{W}}$ and prior $\varphi(w)$. 
    Let $W \subseteq \mathcal{W}$ be the support of $\varphi(w)$ and $W_0 \subseteq W$ be the set of global 
    minima of $L(w)$ restricted to $W$. 
    The \emph{fundamental conditions of SLT} are as follows:
    \begin{enumerate}
        \item For all $w\in W$ the support of $p(x|w)$ is equal to the support of $q(x)$.
        \item The prior's support $W$ is compact with non-empty interior, and can be written as the intersection of finitely many
        analytic inequalities. 
        \item The prior $\varphi(w)$ can be written as $\varphi(w) = \varphi_1(w)\varphi_2(w)$ where $\varphi_1(w)\geq 0$ 
        is analytic and $\varphi_2(w) > 0$ is smooth. 
        \item For all $w_0, w_1 \in W_0$ we have $p(x|w_0) = p(x|w_1)$ almost everywhere. 
        \item Given $w_0 \in W_0$, the function $g(x, w) = \log\frac{p(x|w_0)}{p(x|w)}$ satisfies:
        \begin{enumerate}
            \item For each fixed $w \in W$, $g(x, w)$ is in $L^s(q)$ for $s \geq 2$.
            \item $g(x, w)$ is an analytic function of $w$ which can be analytically extended to a 
            complex analytic function on an open subset of $\C^d$.
            \item There exists $C> 0$ such that for all $w\in W$ we have 
            \[
                \E [g(X, w)] \geq C \E [g(X, w) ^2] \qquad X\sim q(x)
            \]
        \end{enumerate}
    \end{enumerate}
    Conditions 4 and 5c are together called the \emph{relatively finite variance} condition.
\end{definition}

\clearpage

\section{Samplers} \label{sec:samplers}

In this section we provide pseudocode 
for the samplers we benchmark: 
SGLD (\cref{alg:sgld}), AdamSGLD (\cref{alg:adamsgld}), 
RMSPropSGLD (\cref{alg:rmspropsgld}), SGHMC (\cref{alg:sghmc}) and 
SGNHT (\cref{alg:sgnht}).

\begin{algorithm}[h!]
    \caption{SGLD} \label{alg:sgld}
  \begin{algorithmic}[1]

    \STATE {\bfseries Inputs:} $\ve w_0 \in \R^d$ (initial parameter), $D_n = \{\ve z_1, \ldots, \ve z_n\}$ (dataset), $\ell : \R^N \times \R^d \to \R$ (loss function)
    \STATE {\bfseries Outputs:} $\ve w_1, \ldots, \ve w_T \in \R^d$
    \STATE {\bfseries Hyperparameters:} $\epsilon > 0$ (step size), $\gamma > 0$ (localization), $\tilde{\beta} > 0$ (posterior temperature), $m \in \N$ (batch size)
    \STATE %
    
    \FOR{$t \gets 0:T-1$}
        \STATE Draw a batch $z_{u_1}, \ldots, z_{u _m}$ from $D_n$.
        \STATE $\ve g_t \gets \frac{1}{m} \sum _{k=1} ^m \grad _{\ve w} \ell(\ve z_{u_k}, \ve w_t)$ \COMMENT{Gradient with respect to the parameter argument.}
        \STATE Draw $\ve \eta _t$ from the standard normal distribution on $\R^d$.
        \STATE $\Delta \ve w_t \gets \tfrac{-\epsilon}{2} \left(\gamma (\ve w_t - \ve w_0) + \tilde{\beta} \ve g_t \right) + \sqrt{\epsilon} \ve \eta _t$.
        \STATE $\ve w_{t+1} \gets \ve w_t + \Delta \ve w_t$
    \ENDFOR
\end{algorithmic}
\end{algorithm}

\begin{algorithm}[h!]
    \caption{AdamSGLD} \label{alg:adamsgld}
  \begin{algorithmic}[1]

    \STATE {\bfseries Inputs:} $\ve w_0 \in \R^d$ (initial parameter), $D_n = \{\ve z_1, \ldots, \ve z_n\}$ (dataset), $\ell : \R^N \times \R^d \to \R$ (loss function)
    \STATE {\bfseries Outputs:} $\ve w_1, \ldots, \ve w_T \in \R^d$
    \STATE {\bfseries Hyperparameters:} $\epsilon > 0$ (base step size), $\gamma > 0$ (localization), $\tilde{\beta} > 0$ (posterior temperature), $m \in \N$ (batch size), $a > 0$ (stability), $b_1, b_2 \in (0, 1)$ (EMA decay rates)
    \STATE %

    \STATE $\ve m_{-1} \gets (0, 0, \ldots, 0) \in \R^d$.
    \STATE $\ve v_{-1} \gets (1, 1, \ldots, 1) \in \R^d$.
    \FOR{$t \gets 0:T-1$}
        \STATE Draw a batch $z_{u_1}, \ldots, z_{u _m}$ from $D_n$.
        \STATE $\ve g_t \gets \frac{1}{m} \sum _{k=1} ^m \grad _{\ve w} \ell(\ve z_{u_k}, \ve w_t)$ \COMMENT{Gradient with respect to the parameter argument.}
        \STATE $\ve m_t \gets b_1 \ve m_{t-1} + (1-b_1) \ve g_t$
        \STATE Define $\ve v_t$ by $\ve v_t[i] \gets b_2 \ve v_{t-1}[i] + (1 - b_2) \ve g_t[i]^2$ for $i=1, \ldots, d$.
        \STATE $\hat{\ve m}_t \gets \frac{1}{1-b_1^t}\ve m_t$.
        \STATE $\hat{\ve v}_t \gets \frac{1}{1 - b_2^t} \ve v_t$. 
        \STATE Define $\ve \epsilon _t$ by $\ve \epsilon _t[i] \gets \frac{\epsilon}{\sqrt{\hat{\ve v}_t [t]} + a}$ for $i=1, \ldots, d$. \COMMENT{Step size of each parameter.}
        \STATE Draw $\ve \eta _t$ from the standard normal distribution on $\R^d$.
        \STATE Define $\Delta \ve w_t$ by 
        $\Delta \ve w_t[i] \gets \tfrac{-\ve \epsilon_t[i]}{2} \left(\gamma (\ve w_t[i] - \ve w_0[i]) + \tilde{\beta} \hat {\ve m}_t[i] \right) + \sqrt{\ve \epsilon_t[i]} \ve \eta _t[i]$ for $i=1, \ldots, d$.
        \STATE $\ve w_{t+1} \gets \ve w_t + \Delta \ve w_t$
    \ENDFOR
\end{algorithmic}
\end{algorithm}

\begin{algorithm}[h!]
    \caption{RMSPropSGLD} \label{alg:rmspropsgld}
  \begin{algorithmic}[1]

    \STATE {\bfseries Inputs:} $\ve w_0 \in \R^d$ (initial parameter), $D_n = \{\ve z_1, \ldots, \ve z_n\}$ (dataset), $\ell : \R^N \times \R^d \to \R$ (loss function)
    \STATE {\bfseries Outputs:} $\ve w_1, \ldots, \ve w_T \in \R^d$
    \STATE {\bfseries Hyperparameters:} $\epsilon > 0$ (base step size), $\gamma > 0$ (localization), $\tilde{\beta} > 0$ (posterior temperature), $m \in \N$ (batch size), $a > 0$ (stability), $b \in (0, 1)$ (EMA decay rate)
    \STATE %
    
    \STATE $\ve v_{-1} \gets (1, 1, \ldots, 1) \in \R^d$.
    \FOR{$t \gets 0:T-1$}
        \STATE Draw a batch $z_{u_1}, \ldots, z_{u _m}$ from $D_n$.
        \STATE $\ve g_t \gets \frac{1}{m} \sum _{k=1} ^m \grad _{\ve w} \ell(\ve z_{u_k}, \ve w_t)$ \COMMENT{Gradient with respect to the parameter argument.}
        \STATE Define $\ve v_t$ by $\ve v_t[i] \gets b \ve v_{t-1}[i] + (1 - b) \ve g_t[i]^2$ for $i=1, \ldots, d$.
        \STATE $\hat{\ve v}_t \gets \frac{1}{1 - b^t} \ve v_t$
        \STATE Define $\ve \epsilon _t$ by $\ve \epsilon _t[i] \gets \frac{\epsilon}{\sqrt{\hat{\ve v}_t [t]} + a}$ for $i=1, \ldots, d$. \COMMENT{Step size of each parameter.}
        \STATE Draw $\ve \eta _t$ from the standard normal distribution on $\R^d$.
        \STATE Define $\Delta \ve w_t$ by 
        $\Delta \ve w_t[i] \gets \tfrac{-\ve \epsilon_t[i]}{2} \left(\gamma (\ve w_t[i] - \ve w_0[i]) + \tilde{\beta} \ve g_t[i] \right) + \sqrt{\ve \epsilon_t[i]} \ve \eta _t[i]$ for $i=1, \ldots, d$.
        \STATE $\ve w_{t+1} \gets \ve w_t + \Delta \ve w_t$
    \ENDFOR
\end{algorithmic}
\end{algorithm}

\begin{algorithm}[h!]
    \caption{SGHMC} \label{alg:sghmc}
  \begin{algorithmic}[1]

    \STATE {\bfseries Inputs:} $\ve w_0 \in \R^d$ (initial parameter), $D_n = \{\ve z_1, \ldots, \ve z_n\}$ (dataset), $\ell : \R^N \times \R^d \to \R$ (loss function)
    \STATE {\bfseries Outputs:} $\ve w_1, \ldots, \ve w_T \in \R^d$
    \STATE {\bfseries Hyperparameters:} $\epsilon > 0$ (step size), $\gamma > 0$ (localization), $\tilde{\beta} > 0$ (posterior temperature), $m \in \N$ (batch size), $\alpha > 0$ (friction)
    \STATE %
    
    \STATE Draw $\ve p_0$ from a normal distribution on $\R^d$ with mean $\ve 0$ and variance $\epsilon$.
    \FOR{$t \gets 0:T-1$}
        \STATE Draw a batch $z_{u_1}, \ldots, z_{u _m}$ from $D_n$.
        \STATE $\ve g_t \gets \frac{1}{m} \sum _{k=1} ^m \grad _{\ve w} \ell(\ve z_{u_k}, \ve w_t)$ \COMMENT{Gradient with respect to the parameter argument.}
        \STATE Draw $\ve \eta _t$ from the standard normal distribution on $\R^d$.
        \STATE $\Delta \ve p_t \gets \tfrac{-\epsilon}{2} \left(\gamma (\ve w_t - \ve w_0) + \tilde{\beta} \ve g_t \right) - \alpha \ve p_t + \sqrt{2\alpha \epsilon} \eta_t$.
        \STATE $\ve p_{t+1} \gets \ve p_t + \Delta \ve p_t$.
        \STATE $\ve w_{t+1} \gets \ve w _t + \ve p_t$ 
    \ENDFOR
\end{algorithmic}
\end{algorithm}

\begin{algorithm}[h]
    \caption{SGNHT} \label{alg:sgnht}
  \begin{algorithmic}[1]

    \STATE {\bfseries Inputs:} $\ve w_0 \in \R^d$ (initial parameter), $D_n = \{\ve z_1, \ldots, \ve z_n\}$ (dataset), $\ell : \R^N \times \R^d \to \R$ (loss function)
    \STATE {\bfseries Outputs:} $\ve w_1, \ldots, \ve w_T \in \R^d$
    \STATE {\bfseries Hyperparameters:} $\epsilon > 0$ (step size), $\gamma > 0$ (localization), $\tilde{\beta} > 0$ (posterior temperature), $m \in \N$ (batch size), $\alpha_0 > 0$ (initial friction).
    \STATE %
    
    \STATE Draw $\ve p_0$ from a normal distribution on $\R^d$ with mean $\ve 0$ and variance $\epsilon$.
    \FOR{$t \gets 0:T-1$}
        \STATE Draw a batch $z_{u_1}, \ldots, z_{u _m}$ from $D_n$.
        \STATE $\ve g_t \gets \frac{1}{m} \sum _{k=1} ^m \grad _{\ve w} \ell(\ve z_{u_k}, \ve w_t)$ \COMMENT{Gradient with respect to the parameter argument.}
        \STATE Draw $\ve \eta _t$ from the standard normal distribution on $\R^d$.
        \STATE $\Delta \ve p_t \gets \tfrac{-\epsilon}{2} \left(\gamma (\ve w_t - \ve w_0) + \tilde{\beta} \ve g_t \right) - \alpha_t \ve p_t + \sqrt{2\alpha_t \epsilon} \eta_t$.
        \STATE $\ve p_{t+1} \gets \ve p_t + \Delta \ve p_t$.
        \STATE $\alpha _{t+1} \gets \alpha_t + \|p_t\| / d - \epsilon$
        \STATE $\ve w_{t+1} \gets \ve w _t + \ve p_t$ 
    \ENDFOR
\end{algorithmic}
\end{algorithm}

\clearpage

\end{document}